\documentclass[pdflatex,sn-mathphys-num]{sn-jnl}

\usepackage{graphicx}
\usepackage{multirow}
\usepackage{amsmath,amssymb,amsfonts}
\usepackage{amsthm}
\usepackage{mathrsfs}
\usepackage[title]{appendix}
\usepackage{xcolor}
\usepackage{textcomp}
\usepackage{manyfoot}
\usepackage{booktabs}
\usepackage{algorithm}
\usepackage{algorithmicx}
\usepackage{algpseudocode}
\usepackage{listings}
\usepackage{mathtools}
\mathtoolsset{showonlyrefs}
\usepackage{tikz}
\usetikzlibrary{arrows.meta,calc}
\usepackage{enumitem}
\usepackage{tabularx}
    \newcolumntype{Y}{>{\raggedright\arraybackslash}X}
    \newcolumntype{C}{>{\centering\arraybackslash}X}
\usepackage{array}
\usepackage{threeparttable}

\theoremstyle{thmstyleone}
\newtheorem{theorem}{Theorem}  
\newtheorem{maintheorem}{Main Theorem}
\newtheorem{proposition}[theorem]{Proposition} 
\newtheorem{corollary}[theorem]{Corollary}
\newtheorem{assumption}[theorem]{Assumption}
\newtheorem{lemma}[theorem]{Lemma}
\newtheorem{remark}{Remark}

\theoremstyle{thmstylethree}
\newtheorem{definition}{Definition}
\newtheorem{example}{Example}

\usepackage{color}

\begin{document}

\title{Singularity-Induced Local Complexity for High-Dimensional Learning Models}

\author*[1,2]{\fnm{Atsuyoshi} \sur{Muta}}\email{atsuyoshimuta@g.ecc.u-tokyo.ac.jp}

\author[1,2,3]{\fnm{Masaaki} \sur{Imaizumi}}\email{imaizumi@g.ecc.u-tokyo.ac.jp}

\affil*[1]{\orgname{The University of Tokyo}, \orgaddress{\city{Tokyo}, \country{Japan}}}

\affil[2]{\orgname{RIKEN Center for Advanced Intelligence Project}, \orgaddress{\city{Tokyo}, \country{Japan}}}

\affil[3]{\orgname{Kyoto University}, \orgaddress{\city{Kyoto}, \country{Japan}}}

\abstract{
Many overparameterized models that have achieved remarkable success are singular models.
That is, the set of optimal parameters consists of multiple points, or the Hessian of the generalization error at such parameters is degenerate, reflecting redundancy in the model parametrization.
Therefore, understanding the effective dimension and intrinsic model complexity, beyond the apparent dimension of the parameter space, is essential for studying the generalization ability of such overparameterized models.
In this work, we focus on the \emph{real log canonical threshold (RLCT)} and its localized counterpart, the \emph{local RLCT}, which are fundamental quantities for characterizing  the generalization of singular models.
We clarify the theoretical relationship between the local RLCT and existing two key metrics related to generalization under some localization assumption: 
(I) the rank of the Hessian matrix of the generalization error and (II) the covering number of a family of loss functions.
This result provides novel insights into effective dimension and intrinsic complexity of overparameterized models.
}

\keywords{
Overparameterized models,
local model complexity,
singular learning theory,
real log canonical threshold,
covering number,
generalization bound 
}

\maketitle
\section{Introduction}

\subsection{Background}

In machine learning, 
many learning models like neural networks are \textit{overparameterized}, which have more parameters than samples, and are known to achieve good generalization performance on a wide range of tasks \cite{krizhevsky2017imagenet, simonyan2014very, hinton2012deep}. 
In contrast, 
classical learning theory suggests that an excess number of parameters can lead learning models to overfitting and prevents improvement in predictive performance \cite{vapnik1998statistical, bartlett2002rademacher}. 
Therefore, the theoretical mechanisms underlying the generalization of overparameterized models is not yet fully understood, and this is an open problem \cite{zhang2016understanding, Nagarajan2019-uc}.

To tackle the problem, the \textit{singular learning theory} \cite{wei2022deep, lau2023local, watanbe2001nonident, Aoyagi2024-fz} has been utilized to understand overparameterized models through a notion of \textit{model singularity}.
Here, model singularity means that a learning model has either of the two conditions \cite{watanabe2013widely}:
(i) there exist multiple parameters that minimize the generalization error (non-identifiability), or
(ii) the generalization error has a degenerate Hessian at its minimizers (non-regularity).
The singular learning theory shows that, for example, a performacne of Bayesian methods for overparameterized models with model singularity is described by a key quantity named \textit{real log canonical threshold} (RLCT) \cite{watanbe2001nonident, Watanabe2001-sq, watanabe2009algebraic}.
In particular, \cite{Watanabe2001-sq} pointed out that the RLCT behaves as an effective dimension in Bayesian inference and showed that it plays an essential role in generalization.
To broaden its applicability, \cite{lau2023local} introduced the \emph{local RLCT}, also called the \emph{local learning coefficient} (LLC), as a localized version of the RLCT, and empirically demonstrated its relationship with the generalization performance of overparameterized models \cite{wei2022deep, lau2023local, Hoogland2024-xc, Wang2024-xw}.

Despite the development, the studies of the singular learning theory still have some limitations. 
First, the relationship between the local RLCT and generalization performance has so far been established empirically,
and theoretical guarantee of them has not yet been established.
Furthermore, the theoretical validation for the (non-local) RLCT is limited to the Bayesian setting, so its applicability to a broad range of learning algorithms has been restricted.

\subsection{Our study}
In this work, we establish theoretical connections between the local RLCT and certain notions of effective dimension and complexity that are related to good generalization.
We derive upper bounds using the local RLCT for two key metrics in the evaluation of learning models under suitable localization conditions: 
(I) the rank of the Hessian matrix of the generalization error and 
(II) the covering number of a family of loss functions by a learning model. 
These metrics are associated with effective dimension \cite{singh2021, singh2023} and model complexity \cite{bartlett2002rademacher, zhang2016understanding}, respectively, and have both been studied as potential tools for explaining the generalization ability of overparameterized models. 
These results provide a theoretical explanation for the observed relationship between the local RLCT and the generalization error \cite{lau2023local}.

Our contributions can be summarized as follows.
\begin{itemize}
    \item \textit{Hessian rank bound using the local RLCT (Theorem \ref{thm:hessian_rank_ineq}}):
    We show that the local RLCT provides an upper bound on the rank of the Hessian of the generalization error at its local minimizers. Let $R(w)$ denote the generalization error at a parameter $w \in W$, and $w_* \in W$ be a local minimizer of $R$. Then, by local RLCT $\lambda(w_*)$ at $w_*$ satisfies
    \[
     \operatorname{rank} \nabla^2 R(w_*) \le 2 \lambda(w_*).
    \] 
    This result establishes a theoretical connection between existing measure of effective dimension and the local RLCT.
    It has been pointed out that the rank of the Hessian of the generalization error can be interpreted as a local effective dimension \cite{singh2021, singh2023, MacKay1992-qx, Maddox2020-qk}. 
    Indeed, this inequality implies that a smaller local RLCT also leads to a smaller Hessian rank.

    \item \textit{Local covering number bound (Theorem \ref{thm:local_covering_bound})}: 
    We derive an upper bound in terms of the local RLCT on the local covering number around a grobal minimizer of generalization error.
    Informally, we establish a upper bound of the following form: 
    let $w_0$ be a parameter that is a global minimizer of the generalization error, 
    then there exist $C > 0$ and $\rho > 0$ such that for any $\tau > \rho$ 
    \begin{align*}
		N\left(\mathcal{F}_{w_0}, L^{\infty}(P), \tau\right) 
		\leq \left(1 + \frac{C}{\tau - \rho}\right)^{2\lambda (w_0)}.
    \end{align*}
    Here, $\mathcal{F}_{w_0}$ is the function space induced by parameters in a neighborhood of $w_0$ and $\lambda (w_0)$ is the local RLCT.
    Note that standard covering number bounds are typically considered for all $\tau>0$, whereas in this work we only consider the range $\tau>\rho$ for some constant $\rho>0$.
    This is because we are interested in a local bound on the generalization error (see Corollary~\ref{cor:local_gen_bound}).
    This upper bound does \emph{not} explicitly depend on the number of parameters, whereas classical bounds typically do \cite{zhang2016understanding, bartlett2002rademacher}, which can prevent them from providing sharp bound in overparameterized models.

    \item \textit{Connecting theoretical insights from Bayesian inference to other inference methods}:
    Our results provide theoretical support for interpreting the local RLCT as a measure of effective dimension or complexity.
    \cite{Watanabe2001-sq} pointed out that the RLCT behaves like an effective dimension in Bayesian inference. 
    Moreover, \cite{lau2023local} empirically showed a similar interpretation of the local RLCT (LLC) can hold even for models trained by empirical risk minimization rather than Bayesian inference.
    Our result provides a theoretical explanation for this interpretation.
\end{itemize}

We present two technical contributions of our proof to achieve the main results:
First, for the Hessian rank bound, the key idea is to combine the splitting lemma 
with the invariance of the RLCT under diffeomorphisms.
The splitting lemma 
separates the quadratic contribution to the Hessian from the higher-order contribution, and combining this separation with the monotonicity  (Proposition~\ref{lm:rlct}) and invariance (Proposition~\ref{lem:diffeomor_invarient}) of the local RLCT under diffeomorphisms bounds the rank of the Hessian.
Second, for the covering number bound, we consider the covering number after embedding the parameters into a function space through the loss function.
In this case, the local covering number is mainly determined by the second-order contribution of the Hessian near the zeros of the loss function, then we can obtain an upper bound on the covering number.

\subsection{Related Works}

There is a broad range of studies that focus on the geometry of the parameter space of learning models, including singularity, spanning from classical works to those developed after the rise of deep learning. Here, we consider (i) classical influential studies on singularity, (ii) studies that investigate the generalization error of deep learning using flatness or sharpness, and (iii) studies that leverage singularity to analyze generalization error in deep learning. The present work is positioned within (iii).

(i) Classical studies on singularities include \cite{amari2003learning, amari2006singularities, fukumizu1999generalization, watanbe2001nonident, watanabe2010asymptotic}.
In \cite{amari2006singularities}, which investigates multilayer-perceptron learning dynamics, singularities are shown to have a strong effect on the dynamics.
The works \cite{watanbe2001nonident, watanabe2009algebraic} established key results on the asymptotic generalization error of singular models under Bayesian inference, where algebraic-geometric singularities of the zero set of the generalization error  determine the Bayesian generalization error.

(ii) Pioneering studies that investigate the generalization error of deep learning using flatness or sharpness are \cite{hochreiter1997flat}.
Later, \cite{keskar2016large} reported that large-batch training tends to converge to sharp minima and generalizes worse, suggesting a close link between flatness and generalization.
 However, \cite{Dinh2017-wt} showed that such flatness can be changed arbitrarily by reparameterization, so Hessian-eigenvalue-based measures are not invariant. In contrast, the local real log canonical threshold used here is invariant under any diffeomorphic change of variables and can be viewed as a new notion of flatness.

 (iii) Representative studies that leverage singularity to analyze generalization error in deep learning include \cite{wei2022deep, lau2023local, Aoyagi2024-fz, Nagayasu2023-fj}.
The work \cite{wei2022deep} pointed out that singular learning theory has not been effectively incorporated into the theory of deep learning, and aimed to propose potential applications.
\cite{lau2023local} attempts a more concrete application and argues that the local learning coefficient is an appropriate complexity measure for deep models, focusing on empirical validation.
In contrast, this study proves a theoretical inequality relating the Hessian rank and the local real log canonical threshold, and derives a theoretical upper bound.
\cite{Wang2024-xw} pointed out a connection between the LLC (local RLCT) and the Hessian rank briefly in their settings in the footnote of Appendix~D.3. 
In this work, however, we give a rigorous formulation of this relationship and apply it to clarify its relationship with existing complexity measures.
\cite{Aoyagi2024-fz} computes the real log canonical threshold for linear neural networks with identity activation.
Since \cite{Aoyagi2024-fz, Nagayasu2023-fj} focus on Bayesian generalization error, their goals differ from ours; we also contribute by using the real log canonical threshold as a complexity measure.

\subsection{Notation} \label{sec:notation}
Let $k \in \mathbb{N}$ and  $\langle \cdot, \cdot \rangle$ is the standard inner product on $\mathbb{R}^k$, and $\|\cdot\|$ is the standard norm on $\mathbb{R}^k$. 
For a point $x_0 \in \mathbb{R}^k$ and radius $\delta>0$, the open ball $\{x \in \mathbb{R}^k : \|x-x_0\| < \delta\}$ is denoted by $B_{\delta}(x_0)$. 
We write $B_{\delta}^k(x_0)$ when we wish to make the dimension $k$ explicit.
Let $\|\cdot\|_F$ denote the Frobenius norm for a real matrix.
For symmetric matrices $A$ and $B$, we write $A \preceq B$ if $B-A$ is positive semidefinite.
$I_d$ denotes the $d \times d$ identity matrix with $d \in \mathbb{N}$.
For $i, j \in \mathbb{N}$, $I_{i \times j}$ denotes the $i \times j$ matrix whose $(k,k)$-entries are $1$ for $1 \le k \le \min\{i,j\}$ and whose other entries are $0$.
Similarly, $0_{i \times j}$ denotes the $i \times j$ zero matrix.

Let $\mathcal{M} \subset \mathbb{R}^k$ be a Borel set and let $\mu$ be a Borel measure on $\mathcal{M}$. 
For $p \in [1, \infty]$, $\|\cdot\|_p$ denotes the norm of $L^p(\mu) = L^p(\mathcal{M}, \mu)$.
Let $P$ be a Borel probability measure on $\mathcal{M}$, and let $X, X' \sim P$ be random vectors.
We write $\operatorname{Cov}(X, X') := \mathbb{E}\big[(X - \mathbb{E}[X])(X' - \mathbb{E}[X'])^\top\big]$ and $\operatorname{Cov}(X) := \operatorname{Cov}(X, X)$.

\section{Preliminaries}

\subsection{Setup: Learning from Data}\label{sec:basic_setup}

We introduce the standard learning problem.
Let $d \in \mathbb{N}$ denote the number of parameters, and let the parameter space $W \subset \mathbb{R}^d$ be a bounded open set.
Moreover, let $m_1,m_2 \in \mathbb{N}$ denote the input and output dimensions, respectively. 
Let $\mathcal{X} \subset \mathbb{R}^{m_1}$ and $\mathcal{Y} \subset \mathbb{R}^{m_2}$ be compact input and output spaces, respectively. 
Let $\mathcal{Z} = \mathcal{X} \times \mathcal{Y} \subset \mathbb{R}^m$ be the data space with $m=m_1+m_2$.

Let $\phi:\mathcal{X}\times W\to\mathcal{Y}$ be measurable function, and define the parameterized model class $\mathcal{F}=\{\phi_w:\mathcal{X}\to\mathcal{Y}\mid w\in W\}$, where $\phi_w(x)=\phi(x,w)$.
Let $L:\mathcal{Y}\times\mathcal{Y}\to[0,\infty)$ be a nonnegative measurable function.
For $z=(x,y)\in\mathcal{Z}$ and $w\in W$, we define the loss function associated with the model $\phi_w$ by
$\ell(z,w)
=
L\bigl(y,\phi_w(x)\bigr)$.

In this problem, we select parameters based on observed data points and evaluate their performance.
Let $Z=(X,Y)\sim P$ be a random vector, and let $Z_1,\ldots,Z_n$ be i.i.d.\ copies of $Z$.
Then we denote the dataset by $S = (Z_1, \ldots, Z_n)$.
The empirical error and the generalization error are defined as
\[
R_S(w)=\frac{1}{n}\sum_{i=1}^n \ell(Z_i,w), \qquad R(w) = \mathbb{E}_{Z \sim P}[\ell(Z, w)] \qquad (w \in W).
\]

In the typical learning problem, a model parameter $w$ is selected based on this empirical error (for example, the minimizer of the empirical error $R_S(w)$), and its performance is then evaluated using the generalization error $R(w)$ \cite{bartlett2002rademacher, zhang2016understanding, Bousquet2002-jg}.
Since the generalization error cannot be directly observed from the dataset, 
it is essential to control a gap between the empirical error and the generalization error \cite{bartlett2002rademacher, Bousquet2002-jg}.

\paragraph{Assumption and Notation}
We introduce basic assumptions in this work.
\begin{assumption}[Realizability]\label{asmp:realizability}
Let $\mathcal{G}$ denote the class of all measurable functions
from $\mathcal{X}$ to $\mathcal{Y}$.
There exists $w_0 \in W$ such that
\[
    R(w_0)=R^\star :=
    \inf_{g\in\mathcal{G}}
    \mathbb{E}\bigl[L(Y,g(X))\bigr].
\]
\end{assumption}
\begin{assumption} \label{asmp:analytic}
$\ell(z,w)$ is analytic on $\mathcal{O}_{\mathcal{Z}} \times W$ for some open set $\mathcal{O}_{\mathcal{Z}}$
containing $\mathcal{Z}$.
\end{assumption}

Assumption~\ref{asmp:realizability} is often assumed in machine laerning and coincides with the realizability in prior work \cite{Watanabe1999-be, WATANABE201020} (see Remark~\ref{rem:kl_eq}).
Assumption \ref{asmp:analytic} is also natural as discussed in Appendix~\ref{subsec:asmp_validity}.
With this assumption, we can state that the generalization error $R(w)$ is analytic on $W$ (see Proposition \ref{prop:gen_err_analytic} in Section~\ref{subsec:gen_err_analytic}). 

As basic notation, we write $w_0 \in W$ as a global minimzier of the generalization error $R(w)$.
We also define a set of global minimizers of the generalization error as
\begin{align*}
    W_0 &= \operatorname*{arg\,min}_{w \in W} R(w) = \{w \in W :  R(w)=R^\star\}.
\end{align*}
Further, we write $w_* \in W$ as a local minimzier of $R(w)$.

\begin{remark}\label{rem:kl_eq}
    In standard settings, $R(w) - R^\star = 0$ if and only if the KL divergence from the true distribution to the model at $w$ is zero;
    this holds, for example, for multiclass classification with the log loss and for regression with the squared loss under Gaussian noise \cite[Examples~1 and~2]{lahlou2023deup}.
    Hence, Assumption~\ref{asmp:realizability} is equivalent to the realizability condition in \cite{Watanabe1999-be, WATANABE201020} in these settings.
\end{remark}

\subsection{Our Interest: Local Complexity Measures of Learning}\label{subsec:our_interest}

The focus of this study is to evaluate the local complexity measures.
Specifically, we consider the problem of evaluating the following metrics: 
(I) a Hessian matrix of the generalization error at local minimizers, 
and (II) a covering number of models in a neighborhood of the global minimizers.

Below, we define complexity measures of $R(w)$ at the corresponding points.

\paragraph{I. Hessian Rank}
The first measure of interest is the rank of the Hessian of the generalization error $R(w)$. 
In this work, we focus on its value at local minimizers $w_* \in W$:
\begin{align}
\operatorname{rank} \nabla^2 R(w_*).
\label{eq:rank}
\end{align}

We evaluate the Hessian rank for two reasons: (i) it is an intuitively reasonable measure of effective dimension, (ii) it is closely related to classical quantities that have been proposed as effective dimension measures.
Indeed, for (i), the Hessian rank is the number of eigenvectors associated with positive eigenvalues, which is the number of directions along which substantial changes can occur in a neighborhood of $w_*$ \cite{singh2021, singh2023}. Based on this property, previous studies bounded the Hessian rank of deep linear fully connected networks, and deep linear convolutional networks \cite{singh2021}.  
Moreover, for (ii), the Hessian rank is closely related to classical effective dimension that have also been studied in overparameterized models. 
First, it is an upper bound of MacKay’s effective dimension \cite{singh2021, MacKay1992-qx, Maddox2020-qk}; $ \gamma(\varepsilon) = \sum_{i=1}^d \frac{\lambda_i}{\lambda_i + \varepsilon}, \varepsilon > 0$, where $\lambda_i$ denotes the $i$-th eigenvalue of the Hessian.
Since $\gamma(\varepsilon) \le \operatorname{rank} \nabla^2 R(w_*)$  holds, the Hessian rank is an upper bound on MacKay's effective dimension.
Second, this Hessian rank connected to the Fisher information rank at global minimizers, 
which has been pointed out as a measure of effective dimension \cite{Berezniuk2020-ej, Datres2024-xc, Sun2025-qr}.
That is, when the negative log-likelihood loss is considered, the Fisher information matrix $F(w_0)$ satisfies
$F(w_0)=\nabla^2 R(w_0)$ at $w_0 \in W_0$.

\paragraph{II. Local Covering Number}
The second measure of interest is a local covering number of a neighborhood of $w_0 \in W_0$.
In preparation, the covering number is defined as follows:
\begin{definition}[Covering Number] \label{def:covering}
Let $(\mathbb{D}, \rho)$ be a metric space.
A finite set $\{x_1, \ldots, x_N\} \subset \mathbb{D}$ for some $N \in \mathbb{N}$ is called $\tau$-cover of $\mathbb{D}$ with respect to the metric $\rho$ if
$
\mathbb{D}
\subset
\bigcup_{j=1}^{N} \{x \in \mathbb{D} : \rho(x,x_j) \le \tau\}
$
holds.
The covering number $N\left(\mathbb{D}, \rho, \tau\right)$ is the minimum cardinality among all $\tau$-covers of $\mathbb{D}$ with respect to the  metric $\rho$.
\end{definition}

Covering numbers are a standard tool in the analysis of the generalization error \cite{zhang2016understanding, bartlett2002rademacher, wainwright2019high, gine2021mathematical}. 
The importance of this quantity lies in its ability to control the gap between the empirical error and the generalization error \cite{wainwright2019high, gine2021mathematical}; see also Proposition~\ref{prop:syutyu_ineq}.

With the notion, our interest is a covering number of a local set of models $\mathcal{F}_{\ell}(B_{\delta}(w_0)) = \{\ell(\cdot, w) : w \in B_{\delta}(w_0)\}$ 
with $\delta > 0$:
\begin{align}    
N\left(\mathcal{F}_{\ell}(B_{\delta}(w_0)), L^{\infty}(P), \tau\right).  \label{eq:covering}
\end{align}
Of particular interest is the case in which $\mathbb{D}=\mathcal{F}_{\ell}(A)$ and $d$ is the distance induced by the norm $\|\cdot\|_{\infty}$.
Since the metric is clear from the context, we simply write $L^{\infty}(P)$ instead of $\|\cdot\|_{\infty}$.

The covering number of such local sets is important for analyzing modern machine learning methods, such as deep learning. Since learning algorithms restrict parameters to specific local sets, the covering number of these sets is examined to characterize generalization error \cite{Neyshabur2015-er,Bartlett2017-gi,Golowich2018-sj}.

\subsection{Local Real Log Canonical Threshold (Local RLCT)}

We introduce the notion of the \textit{real log canonial threthold} (RLCT) and the \textit{local real log canonial threthold} (local RLCT) as basic concepts for this study.

\paragraph{RLCT}

\begin{figure}[t]
  \centering  
  \begin{minipage}{0.55\linewidth}
    \centering
    \resizebox{1\columnwidth}{!}{
        \begin{tikzpicture}[font=\footnotesize, line cap=round]
            \def\pad{0.25}
            \def\yRe{-0.30}
            \def\xIm{1.80}

            \def\xA{-2.0}
            \def\xB{-1.2}
            \def\xC{-0.2}
            \def\xL{ 1.0}

            \def\hA{0.0}
            \def\hB{0.0}
            \def\hC{0.0}
            \def\hL{0.0}

            \def\rPole{0.06}

            \coordinate (zetaPos) at (-0.2, 1.10);

            \coordinate (apexPos) at (-0.55, 0.65);

            \tikzset{
              darr/.style={densely dotted, ->, line width=0.6pt}
            }

            \path[use as bounding box]
              (-2.3-\pad, -1.3-\pad) rectangle (2.7+\pad, 1.3+\pad);

            \draw[->] (-2.3,\yRe) -- (2.7,\yRe) node[below left] {Re};
            \draw[->] (\xIm,-1.3) -- (\xIm,1.3) node[above] {Im};

            \node[below left] at (\xIm,\yRe) {$0$};

            \coordinate (pA) at (\xA,{ \yRe+\hA });
            \coordinate (pB) at (\xB,{ \yRe+\hB });
            \coordinate (pC) at (\xC,{ \yRe+\hC });
            \coordinate (pL) at (\xL,{ \yRe+\hL });

            \draw (pA) circle[radius=\rPole];
            \draw (pB) circle[radius=\rPole];
            \draw (pC) circle[radius=\rPole];
            \filldraw[fill=red!80!black] (pL) circle[radius=\rPole];

            \node[above] at ($(pL)+(0,0.2)$) {$-\lambda$};

            \node at (zetaPos) {$\displaystyle \zeta(s) = \int_{C} (R(w) - R(w_{\mathrm{min}}))^s dw.$};

            \draw[darr,-{Stealth[length=1.5mm,width=1.5mm]},opacity=0.7] (apexPos) -- ($(pA)+(0,0.1)$);
            \draw[darr,-{Stealth[length=1.5mm,width=1.5mm]},opacity=0.7] (apexPos) -- ($(pB)+(0,0.1)$);
            \draw[darr,-{Stealth[length=1.5mm,width=1.5mm]},opacity=0.7] (apexPos) -- ($(pC)+(0,0.1)$);
            \draw[darr,-{Stealth[length=1.5mm,width=1.5mm]},opacity=0.7] (apexPos) -- ($(pL)+(0,0.1)$);

            \node[below, font=\tiny, align=center, draw] at (-0.4,-0.6) {
                The poles of $\zeta(s)$ are aligned on \\
                the negative real axis.
            };

        \end{tikzpicture}
    }
    {\small \textbf{(a) The definition of RLCT.} }
  \end{minipage}
   \hspace{-0.01\linewidth}
  \begin{minipage}{0.40\linewidth}
    \centering
    \includegraphics[width=0.9\linewidth]{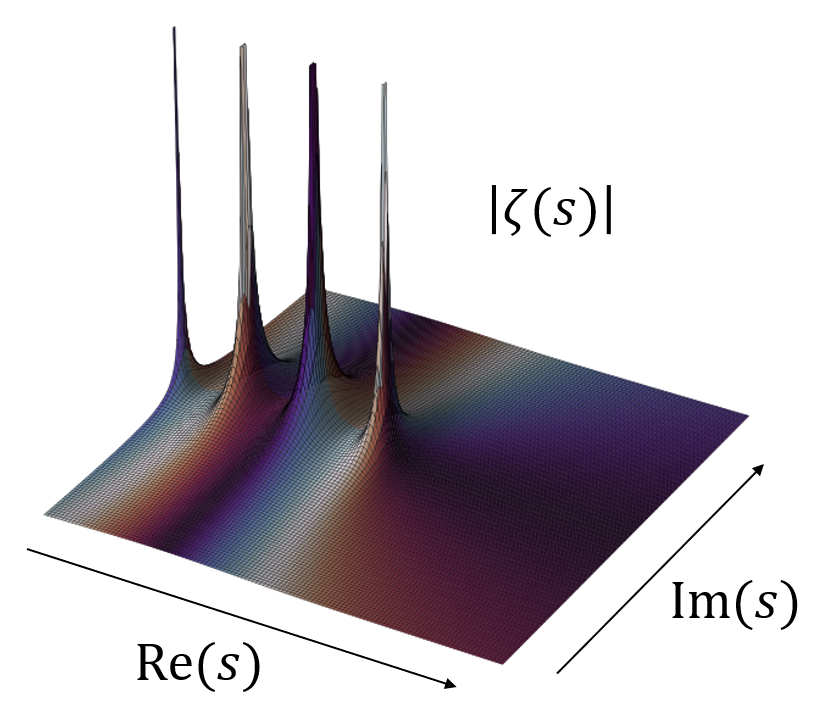}
    {\small \textbf{(b) The image of $\zeta(s)$.} }
  \end{minipage}
  \caption{
        \textbf{(a) The definition of RLCT.}
        The RLCT is defined as the absolute value of the largest pole of the $\zeta$ function.
        The $\zeta$  function is initially defined for $s \in \mathbb{C}$ with $\operatorname{Re}(s) > 0$. 
        and it can then be extended to the entire complex plane by analytic continuation \cite{Atiyah1970-et}. 
        Then their poles (Remark~\ref{rem:pole_order}) arise on the negative real axis \cite{kashiwara1976b, watanabe2009algebraic}.
        It was found that this mathematical quantity is deeply related to statistical models \citep{watanabe2009algebraic}.
        \textbf{(b) The image of $\zeta(s)$.}
        At each pole, $|\zeta(s)|$ diverges to infinity.
    }
  \label{fig:rlct}
\end{figure}

We first introduce the RLCT, which is necessary to define the local RLCT.
\begin{definition}{(Real Log Canonical Threshold; RLCT)}
    Let $R$ be analytic on $W$, as ensured by Assumption~\ref{asmp:analytic}. 
    Let $C \subset W$ be a compact set, and $w_{\mathrm{min}} = \operatorname*{arg\,min}_{w \in C} R(w)$. 
    We define its $\zeta$ function of $R(w) - R(w_{\mathrm{min}})$ by
    \[
        \zeta(s) = \int_C \left( R(w) - R(w_{\mathrm{min}}) \right)^s dw \quad (\mathrm{Re}(s) > 0,   s \in \mathbb{C}),
    \]
    Then $\zeta$ admits an analytic continuation to a meromorphic function on $\mathbb{C}$, and all of its poles lie on the negative real axis \cite{Atiyah1970-et} \cite[Corollary~5.2]{kashiwara1976b} \cite[Theorem~2.4]{watanabe2009algebraic}.
    
    We denote by $\lambda(R,C)$ the absolute value of the largest pole (see Remark~\ref{rem:pole_order}) of $\zeta$, and call it the \textit{real log canonical threshold} (RLCT) of $R$ on $C$. 
    When it is clear from the context, we write $\lambda(R,C)$ simply as $\lambda$ (Fig.~\ref{fig:rlct}). 
\end{definition}

\begin{remark}[Poles of complex functions]\label{rem:pole_order}
Poles are notions from complex analysis.
Let $s \in \mathbb{C}$ and $f(s)$ be a complex valued function defined on a neighborhood of $s_0 \in \mathbb C$.
We say that $s_0$ is a pole of $f$ if there exist an open neighborhood $U$ of $s_0$ and a natural number $m>0$ such that 
\[
    f(s)=\frac{g(s)}{(s-s_0)^m} \quad (s \in U \setminus \{s_0\}),
\]
where a complex function $g(s)$ that is differentiable on $U$ with $g(s_0) \neq 0$.
\end{remark}

The RLCT was first introduced in singular learning theory to analyze a key quantity characterizing singular models \cite{Watanabe2001-kc, aoyagi2005stochastic, Watanabe2006-bo, watanabe2010asymptotic}.
In the original formulation, the RLCT $\lambda(R, W)$ was used to analyze the generalization error of Bayesian inference, where $W$ is the whole (compact) parameter space \cite{Watanabe2001-kc},
which suggested that the RLCT behaves as a certain notion of effective complexity.

\paragraph{Local RLCT}

We introduce the local RLCT at local minimizer $w_* \in W$ of the generalization error.
Applications of the RLCT to generalization error analysis have primarily focused on Bayesian inference, while its localized version, the local RLCT (LLC), has recently been shown to play an important role also for models trained by empirical risk minimization \cite{wei2022deep, lau2023local, Hoogland2024-xc}.

\begin{definition}{(Local RLCT)} \label{def:local_rlct}
    Let $w_* \in W$ be a local minimizer of the generalization error $R$. The limit $\lim_{\delta \to 0} \lambda \left(R, \overline{B_{\delta}}(w_*)\right)$ is called  the local RLCT at $w_*$, and is denoted by $\lambda(R, w_*)$. When $R$ is clear from the context, we write it simply as $\lambda(w_*)$. 
\end{definition}

Note that the limit $\lim_{\delta \to 0} \lambda \left(R, \overline{B_{\delta}}(w_*)\right)$ always exists  (Appendix~\ref{appendix:slt}, Lemma~\ref{lem:def_of_LRLCT}).

\section{Main Results}

We present main results of this study: we first bound the rank of the Hessian \eqref{eq:rank}, and  we then derive an upper bound on the covering number \eqref{eq:covering}.

\subsection{Hessian Rank Bound}

We first show that the local RLCT gives an upper bound on the rank of the Hessian at the point. A underlying idea was previously sketched in \cite[Appendix~D.3]{Wang2024-xw}; here, we formulate the result rigorously.

\begin{maintheorem}[The Hessian Rank Bound]\label{thm:hessian_rank_ineq}
    Under Assumption~\ref{asmp:analytic}, 
    for any $w_* \in W$ that is a local minimizer of $R$, we have
    \begin{align*}        
    \operatorname{rank}   \nabla^2 R(w_*) \leq 2 \lambda(w_*).
        \end{align*}
\end{maintheorem}

This result shows that the local RLCT is an upper bound on one measure of effective dimension at any local minimizer of the generalization error.
Since the Hessian rank can be regarded as a measure of effective dimension as discussed in Section~\ref{subsec:our_interest}, and a small local RLCT necessarily implies a small Hessian rank.

Furthermore, we can confirm that a small local RLCT is related to better generalization through a reduction in the effective dimension.
A more detailed discussion is provided in Theorem~\ref{thm:local_covering_bound} and its application, Corollary~\ref{cor:local_gen_bound}.

\paragraph{Proof Outline}
To prove this theorem, we introduce a fundamental lemma from the singularity theory
Its proof is given in Appendix~\ref{sec:prf_spliting}.

\begin{lemma}[Splitting lemma]\label{lem:splitting}
    Let $R$ be a nonnegative $C^3$ function on $W$, and  $r=\operatorname{rank} \nabla^2 R(w_*)$. 
    Then for any local minimizer $w_* \in W$ there exists a local $C^1$ diffeomorphism $\Phi$ around $w_*$ such that
    $\Phi : \mathbb{R}^r \times \mathbb{R}^{d-r} \to \mathbb{R}^d;  (u,v) \mapsto w$ with $w_*$ corresponding to $(u_*,v_*)$ and
    \[
        R \circ \Phi(u, v)
        =
        \|u\|^2 + \mathrm{Res}(v).
    \]
    Here, ${\mathrm{Res}}$ is a nonnegative $C^2$ function and satisfies
    \[
        {\mathrm{Res}}(v_*)=R(w_*),
        \qquad
        \nabla \mathrm{Res}(v_*)=0,
        \qquad
        \nabla^2 {\mathrm{Res}}(v_*)=0.
    \]
\end{lemma}

The splitting lemma states that there exists a local change of coordinates that separates the function into a canonical quadratic form and a residual higher-order terms.
This allows us to separate the directions along which $R$ can vary substantially near a local minimum and to investigate their relationship with the local RLCT.

Note that, when $R$ is analytic, the coordinate transformation can be chosen analytic, and analytic version is used in the proof of Main Theorem~\ref{thm:hessian_rank_ineq}.

\begin{proof}[Proof of Main Theorem~\ref{thm:hessian_rank_ineq}]
For simplicity, we may assume $R(w_*)=0$ without loss of generality.
By Lemma~\ref{lem:splitting}, there exist neighborhoods $\mathcal{O}_{u,v}, \mathcal{O}_{w}$ of the origin and an analytic diffeomorphism $\Phi:\mathcal{O}_{u,v} \to \mathcal{O}_{w}$ such that
$w = \Phi(u,v)$ for $(u,v) \in \mathcal{O}_{u,v}$ and
 \[
    R \circ \Phi(u, v)
    =
    \|u\|^2 + \mathrm{Res}(v).
 \]
By the definition of the local RLCT, we may take $\delta > 0$ small enough so that $B_\delta(w_*) \subset \mathcal{O}_{w}$. Hence, by the invariance of the RLCT under diffeomorphisms (Proposition~\ref{lem:diffeomor_invarient}),  we have
\begin{align*}
    \lambda \left(R, \overline{B_{\delta}}(w_*)\right) 
    &= \lambda\left(R \circ \Phi, \Phi^{-1} \left(\overline{B_{\delta}}(w_*)\right)\right) \\
    &= \lambda \left(\|u\|^2 + \mathrm{Res}(v), \Phi^{-1}\left(\overline{B_{\delta}}(w_*)\right)\right) \\
    &\geq \lambda \left(\|u\|^2, \Phi^{-1} \left(\overline{B_{\delta}}(w_*)\right)\right) \\
    &= \frac{r}{2}.
\end{align*}
where the inequality uses Proposition~\ref{lm:rlct} and the fact that $\mathrm{Res}(v) \geq 0$.
The last equality follows from Proposition~\ref{lem:rlct_of_squre}.
\end{proof}

\subsection{Local Covering Bound}

We next state the main result, the bound of the covering number in terms of the local RLCT.
Here, we clarify loss functions we consider. We provide some examples after presenting the assumption.
\begin{assumption} \label{asmp:fim}
The loss function $\ell$ satisfies the following condition: there exists a constant $\alpha > 0$ such that for any $w_0 \in W_0$ 
\[ 
\mathrm{Cov} \left(\nabla_w \ell(Z,w_0) \right)
\preceq \alpha \, \nabla^2 R(w_0).
\]
\end{assumption}

The negative log-likelihood loss is a standard example satisfying this condition, and thus multiclass classification with cross-entropy loss and squared loss regression with Gaussian noise satisfy this assumption \cite[Section~4.3.4]{bishop2006pattern} \cite[Section~5.5]{Goodfellow-et-al-2016}.
Since $\mathrm{Cov}\!\left(\nabla_w \ell(Z,w)\right)$ coincides with the Fisher information matrix at every $w_0 \in W_0$, note that $\mathrm{Cov} \left(\nabla_w \ell(Z,w_0) \right) =  \nabla^2 R(w_0)$ holds
\cite{watanabe2009algebraic, watanabe2013widely}. 
Thus, this assumption holds with $\alpha = 1$ for the negative log-likelihood loss.
Another important example is the squared loss regression with bounded outputs. This setup also satisfies this condition (Appendix~\ref{sec:reg_asmp1}).

We give an upper bound of a covering number of a functional set $\mathcal{F}_{\ell}(B_{\delta}(w_0))$  around the global minima $w_0 \in W$.
Here, we provide an upper bound for covering numbers with radius $\tau > \rho_{\delta}$ that exceed a certain threshold $\rho_{\delta} > 0$. Although this upper bound is restrictive than the conventional upper bound for covering numbers with positive $\tau > 0$, this setting is necessary for evaluation using local RLCT.

\begin{maintheorem}[The Local Covering Bound]\label{thm:local_covering_bound}
    Under Assumption~\ref{asmp:realizability}, \ref{asmp:analytic}, \ref{asmp:fim},
    for any $w_0 \in W_0$ there exist constants $L, C >0$ and $\delta > 0$ such that,
    with $\rho_{\delta} = C \delta^2$, the following holds for every $\tau > \rho_{\delta}$:
    \begin{align}\label{ineq:covering_bound_rlct}
		N\left(\mathcal{F}_{\ell}(B_{\delta}(w_0)), L^{\infty}(P), \tau\right) 
		\leq \left(1 + \frac{2 L}{\tau - \rho_{\delta}}\right)^{2\lambda (w_0)}.
    \end{align}
\end{maintheorem}

This result provides several insights:
(i) the small local RLCT reduces the localized complexity of the model around the global minimizer $w_0$, 
(ii) this upper bound improves the previous upper bound for covering numbers based on the parameter dimension $d$ \cite{bartlett2002rademacher, zhang2016understanding},

\paragraph{Proof}
We provide the complete proof here. Main Theorem \ref{thm:hessian_rank_ineq} makes a key contribution for this proof.
\begin{proof}[Proof of Main Theorem~\ref{thm:local_covering_bound}]
    Let $r = \operatorname{rank}   \nabla_w^2 R(w_0)$.
    The case $r=0$ is trivial, so assume $r>0$.
    By the Taylor expansion around $w_0$,
	\[R(w) =
		\frac{1}{2} \left\langle \nabla^2 R(w_0)(w-w_0), (w-w_0)\right\rangle 
		+ o(\|w - w_0\|^2).
	\]
	Then for any $\varepsilon > 0$ there exists $\delta> 0$ such that
	\[\left| R(w) - \frac{1}{2} \left\langle \nabla^2 R(w_0)(w-w_0), (w-w_0)\right\rangle \right| < \frac{\varepsilon}{4}\]
	holds on $B_{\delta}(w_0)$.
    This implies $\frac{1}{2} \left\langle \nabla^2 R(w_0)(w-w_0), (w-w_0)\right\rangle < R(w) + \frac{\varepsilon}{4}$.
    Here, let $\Psi$ be a linear transformation that diagonalizes $\nabla^2 R(w_0)$, and let $u \in \mathbb{R}^r$ denote the directions corresponding to positive eigenvalues, and let $h \in \mathbb{R}^{d-r}$ denote the directions corresponding to zero eigenvalues. Suppose that $w_0$ corresponds to $(u_0, v_0)$ under this transformation. Then
    \begin{align}\label{eq:distance}
        \frac{1}{2} \sum_{i=1}^{r} \lambda_i^+(w_0) (u_i - u_{0,i})^2
        &= \frac{1}{2} \left\langle \nabla^2 R(w_0)(w-w_0), w-w_0 \right\rangle
        < R(w) + \frac{\varepsilon}{4},
    \end{align}
    where $\lambda_i^+(w_0)$ denotes the $i$-th positive eigenvalue.
    Since we may take $\delta$ smaller if necessary, we can assume that
    $R(w) < \frac{\varepsilon}{4}$.
    Then we have
    \begin{align}
		\frac{1}{2} \sum_{i=1}^{r} \lambda_i^+(w_0) (u_i - u_{0,i})^2
		&< \frac{\varepsilon}{2}.
	\end{align}
    This shows that, in the sense of the metric induced by the positive definite matrix determined by the eigendirections corresponding to the nondegenerate eigenvalues of $\nabla^2 R(w_0)$, the radius is less than $\sqrt{\varepsilon}$.

    Then by Lemma~\ref{lem:loss_difference_bound}, 
    there exist $C > 0$ such that
    \begin{align*}
        \bigl\| \ell(\cdot, \Psi(u_1, v_1)) - \ell(\cdot, \Psi(u_2, v_2)) \bigr\|_{\infty}
        &\le L \|u_1-u_2\| + C \delta^2.
    \end{align*}
    Hence, let $B^{r}_{\sqrt{\varepsilon}}(u_0)$ denote the
    $r$-dimensional ball centered at $u_0$, where
    $(u_0,v_0)=\Psi^{-1}(w_0)$, with radius $\sqrt{\varepsilon}$. 
    Then, letting $\rho_{\delta} = C \delta^2$, for any $\tau>\rho_{\delta}$
    \begin{align*}
        N(\mathcal{F}_{\ell}(B_{\delta}(w_0)), L^{\infty}(P), \tau)
        &\leq
        N \left(
            B^{r}_{\sqrt{\varepsilon}}(u_0),
            \|\cdot\|,
            \frac{\tau-\rho_{\delta}}{L}
        \right) \\
        &\leq
        \left(
            1
            + \sqrt{\frac{\varepsilon}{\lambda^{+}_{\mathrm{min}}(w_0)}}
            \frac{2L}{\tau-\rho_{\delta}}
        \right)^{r},
    \end{align*}
    where $\lambda^{+}_{\mathrm{min}}(w_0)$ denotes the smallest eigenvalue.
    The second inequality uses the fact that for the distance $d_G(u, u') = \sqrt{\langle G(u-u'), u-u'\rangle}$ induced by a positive definite matrix $G$, if $\lambda_{\mathrm{min}}$ denotes the minimum eigenvalue of $G$, then
    $\lambda_{\mathrm{min}} \langle u-u', u-u'\rangle \leq \langle G(u-u'), u-u'\rangle$ holds.
    That is, for the usual distance $d(u, u') = \|u-u'\|$,
    \[d(u,u') \leq \frac{1}{\sqrt{\lambda_{\mathrm{min}}}} d_G(u,u')\]
    holds, and hence $B^{r}_{\sqrt{\varepsilon}}(u_0)$ is contained in the ball of radius $\sqrt{\varepsilon / \lambda^{+}_{\mathrm{min}}(w_0)}$ under the usual distance.
    
    Since Main Theorem~\ref{thm:hessian_rank_ineq} gives $r\le 2\lambda(w_0)$, the exponent $r$ can be replaced by $2\lambda(w_0)$.
    Finally, since $\varepsilon$ was arbitrary, setting $\varepsilon = \lambda^{+}_{\mathrm{min}}(w_0)$ completes the proof.
\end{proof}

\section{Application}\label{sec:application}

We discuss the implications of the preceding results for generalization error.

\subsection{Local Generalization Error Bounds and Its Implications}

We derive local generalization bounds and discuss the following implication: 
the generalization bound can vary depending on the local RLCT.
We first establish the following corollary, whose proof is given in Appendix~\ref{sec:local_gen_bound}.

\begin{corollary}\label{cor:local_gen_bound}
Under Assumption~\ref{asmp:realizability}, \ref{asmp:analytic}, \ref{asmp:fim}, for every $w_0 \in W_0$ there exist $\delta > 0$ and $L, B, C > 0$ such that, 
for any $\eta > 0$, the following holds with probability at least $1 - \exp\left(-\frac{n\eta^2}{2B^2}\right)$:
\[
    R(w)
    \leq
    R_S(w) + 96 \sqrt{\frac{2 \lambda(w_0) L B}{n}} + C \delta^2 + \eta, \qquad (w \in B_{\delta}(w_0)), 
\]
where $B$ is a constant with $\ell(z,w) \le B$ on $\mathcal Z \times B_{\delta}(w_0)$.
\end{corollary}

This result shows that the generalization error has the smaller upper bound
in neighborhoods of global minimizers with smaller local RLCTs.

Furthermore, since this result holds for every $w \in B_{\delta}(w_0)$, the generalization bound applies to a learned parameter whenever it lies in this neighborhood, even if it depends on the data. We present a natural result with a learned parameter $\hat{w}$.

\begin{corollary}
    Let $w_0 \in W_0$ and $\delta > 0$ be as in Corollary~\ref{cor:local_gen_bound}.
    Suppose  Assumption~\ref{asmp:realizability}, \ref{asmp:analytic}, \ref{asmp:fim} holds with 
    the same notation as in Corollary~\ref{cor:local_gen_bound}.
    Moreover, let $\hat{w}=\hat{w}(S)$ be a $W$-valued random variable depending on the dataset $S$ and asuume that $\hat{w} \in B_{\delta}(w_0)$ holds.
    Then, 
    we have the following bound 
    with probability at least $1 - \exp \left(-\frac{n\eta^2}{2B^2}\right)$:
    \[
        R(\hat{w}) \leq R_S(\hat{w}) + 96 \sqrt{\frac{2 \lambda(w_0) LB}{n}} + C \delta^2 + \eta.
    \]
\end{corollary}

This result shows that Corollary~\ref{cor:local_gen_bound} also applies to a data-dependent parameter $\hat{w}$ provided that $\hat{w} \in B_{\delta}(w_0)$. 
That is, the generalization error bound is smaller when the learned parameter lies in a neighborhood of a global minimizer with a smaller local RLCT.

This result is consistent with several empirical observations.
\cite{lau2023local} indicates that, among solutions selected by empirical risk minimization, those with smaller local RLCTs tend to generalize better.
Moreover, \cite{smith2018bayesian} indicates that larger local Bayesian evidence, which corresponds to a smaller local RLCT, tends to yield better generalization (see Section~3 and Appendix~A).
Our result provides theoretical support for these empirical observations in the special case where the learned parameter lies near a global minimizer.

\subsection{Local Generalization Error Analysis for Reduced-Rank Regression}\label{subsec:rrr}

We derive concrete upper bounds on the generalization error for reduced-rank regression. 
These results show that even among global minimizers, different local RLCTs can lead to different generalization bounds.

\paragraph{Problem Setup}

We introduce the following reduced-rank regression model.
Fix an integer $h$ satisfying $1\le h<\min\{m_1,m_2\}$, and the reduced-rank regression model with an input $x \in [-1,1]^{m_1}$ is given by
\[
\phi(x,w)=U Vx,
\]
where
$U \in \mathbb{R}^{m_2\times h}$,
$V \in \mathbb{R}^{h\times m_1}$, 
$w=(U, V)$, and the parameter space is  $W=(-2,2)^{m_2 \times h} \times (-2,2)^{h \times m_1}$.
The number of parameters is $d=h(m_1+m_2)$.

We consider a noiseless regression problem with the reduced-rank regression model. 
Let $(X,Y)$ be an input-output pair, where $X$ takes values in $[-1,1]^{m_1}$ and $Y$ takes values in a compact subset of $\mathbb{R}^{m_2}$, and it follows the data-generating model 
\[
Y=T^\star X, \quad 
T^\star =
\begin{pmatrix}
I_r & 0\\
0 & 0
\end{pmatrix} \in \mathbb{R}^{m_2\times m_1},
\]
with $r \in \mathbb{N}$.
Moreover, assume that $\mathbb{E}[XX^\top]$ is positive definite.
The loss function is the squared loss
$\ell((X,Y),w)=\|Y-\phi(X,w)\|^2$.
Then, the generalization error is given by
$R(w)
=\mathbb{E}[\|Y-\phi(X,w)\|^2]
=\mathbb{E}[\|T^\star X - U V X\|^2]$.

\paragraph{Derivation of the Local RLCT}

We investigate local RLCTs for the following three global minimizers $w_i = (U_i, V_i) \in \mathbb{R}^{m_2 \times h} \times \mathbb{R}^{h \times m_1}, i = 1,2,3$, that is, those that satisfy $U_i V_i = T^\star$. While  there are more global minizers, we select $w_i, i = 1,2,3$ with representative structures from among them. 
Specifically, we consider 
\begin{align} \label{eq:def_w_rrr}
    &(U_1, V_1) = \left(\begin{pmatrix}
            I_h \\
            0
        \end{pmatrix},\begin{pmatrix}
            I_r & 0 \\
            0 & 0
        \end{pmatrix}
        \right), 
        (U_2, V_2) = \left( 
        \begin{pmatrix}
            I_r & 0 \\
            0 & 0
        \end{pmatrix},
        \begin{pmatrix}
            I_h & 0
        \end{pmatrix}
        \right), \\
        &(U_3, V_3) = \left( 
        \begin{pmatrix}
            I_r & 0 \\
            0 & 0
        \end{pmatrix},
        \begin{pmatrix}
            I_r & 0 \\
            0 & 0
        \end{pmatrix}
        \right),
\end{align}
and then we have $R(w_i)=0  \; (i=1,2,3).$
Their local RLCTs are computed as
\begin{equation}\label{eq:rrr-rlct}
\begin{gathered}
    \lambda(w_1)=\frac{h m_1 + r(m_2 - h)}{2},
    \qquad
    \lambda(w_2)=\frac{h m_2 + r(m_1 - h)}{2}, \\
    \lambda(w_3)
    =
    \frac{1}{2}
    \left\{ r(m_1+m_2-r) + \kappa \right\},
\end{gathered}
\end{equation}
where 
$
\kappa = \min_{\substack{0 \le j \le h-r, j \in \mathbb{Z}}} \left\{j(m_2-r)+(m_1-r-j)(h-r-j)\right\}.
$
This derivation demonstrates that different global minimziers have different local RLCTs.
The detailed derivation is given in Appendix \ref{appendix:local_rlct_RRR}.
In Section~\ref{sec:experiments}, we empirically verify that the inequality in the main theorem holds.

\paragraph{Derivation of the Local Generalization Error}

Finally, we derive the bound on the local generalization error of the reduced rank regression problem.
By selecting several constants in Lemma~\ref{lem:loss_difference_bound},  Corollary~\ref{cor:local_gen_bound} provides that there exist $\delta > 0$ and the following bounds
\begin{align*}
    R(w) &\leq R_S(w) + 96 \sqrt{\frac{\left(h m_1 + r(m_2 - h) \right) L B}{n}} + C \delta^2 + \eta \quad \left(w \in B_{\delta}(w_1)\right), \\
    R(w) &\leq R_S(w) + 96 \sqrt{\frac{\left(h m_2 + r(m_1 - h) \right) L B}{n}} + C \delta^2 + \eta \quad \left(w \in B_{\delta}(w_2)\right), \\
    R(w) &\leq R_S(w) + 96 \sqrt{\frac{\left(r(m_1+m_2-r) + \kappa \right) L B}{n}} + C \delta^2 + \eta \quad \left(w \in B_{\delta}(w_3)\right), \\
\end{align*}
A key insight is that different local geometries yield different bounds, whereas existing studies give the same bound for every parameter.
The complexity term in \cite[Corollary 6]{bunea2011optimal} is controlled by a parameter-independent quantity, which is independent of local poperties with a learned parameter. 
In contrast, our bound varies with the local RLCT at each global minimizer, giving tighter values near minimizers with fewer effective degrees of freedom.
Thus, we obtain a bound that adapts to the local geometry at each parameter.

\subsection{Local Generalization Error Analysis for Linear Neural Network}\label{subsec:lnn}

We derive upper bounds on the generalization error for a linear neural network.

\paragraph{Problem Setup}

We consider the following linear neural network model.
We define
\[
\phi(x, w) = W_{\mathrm{out}} W_{\mathrm{hid}} W_{\mathrm{in}} x,
\]
where $W_{\mathrm{in}} \in \mathbb{R}^{h_1\times m_1}, W_{\mathrm{hid}} \in \mathbb{R}^{h_2\times h_1}, W_{\mathrm{out}} \in \mathbb{R}^{m_2\times h_2}$ and $w = (W_{\mathrm{in}}, W_{\mathrm{hid}}, W_{\mathrm{out}})$.
Then the parameter space is $W = (-2, 2)^{h_1 \times m_1} \times (-2, 2)^{h_2 \times h_1} \times (-2, 2)^{m_2 \times h_2}$.
For simplicity, we asuume $m_1 > h_1 > h_2 > m_2$.

We consider a noiseless regression problem with the realizable true function $Y=0$ almost surely.
This setting is introduced only to simplify the calculation of the local RLCT; essentially the same proof works when $X$ and $Y$ are correlated, as in reduced-rank regression, with a more involved calculation.
Moreover, assume that $\mathbb{E}[XX^\top]$ is positive definite. 
The loss function is the squared loss
$\ell((X,Y),w)=\|\phi(X,w)\|^2$.
Then, the generalization error is given by
$R(w) =\mathbb{E}[\|\phi(X,w)\|^2]$.

\paragraph{Derivation of the Local RLCT}

We investigate local RLCTs for the following three global minimizers $w_i \in \mathbb{R}^{h_2 \times m_1} \times \mathbb{R}^{h_2 \times m_1} \times \mathbb{R}^{m_2 \times h_2}, i = 1,2,3$. As in the previous section, while  there are more global minizers, we select the parameters $w_i, i = 1,2,3$ with representative structures from among them. 
Specifically, let $w_1, w_2, w_3$ be the parameters defined by
\begin{align}\label{eq:def_w_lnn}
    w_1 &= (I_{h_1 \times m_1}, I_{h_2 \times h_1}, 0_{m_2 \times h_2}), \notag\\
    w_2 &= (I_{h_1 \times m_1}, 0_{h_2 \times h_1}, I_{m_2 \times h_2}), \\
    w_3 &= (0_{h_1 \times m_1}, I_{h_2 \times h_1}, I_{m_2 \times h_2}), \notag
\end{align}
and then we have $R(w_i)=0  \; (i=1,2,3).$
Then
\begin{align}\label{eq:lnn-rlct}
\lambda(w_1) = \frac{m_2 h_2}{2}, \qquad 
\lambda(w_2) = \frac{m_2 h_1}{2}, \qquad
\lambda(w_3) = \frac{m_2 m_1}{2}.
\end{align}
The detailed derivation is given in Appendix \ref{appendix:local_rlct_LNN}.
In Section~\ref{sec:experiments}, we empirically verify that the inequality in the main theorem holds.

\paragraph{Derivation of the Local Generalization Error}

We derive the bound on the local generalization error.
By Corollary~\ref{cor:local_gen_bound}, there exist $\delta > 0$ and such that
\begin{align*}
    R(w) &\leq R_S(w) + 96 \sqrt{\frac{m_2 h_2 LB}{n}} + C \delta^2 + \eta \quad \left(w \in B_{\delta}(w_1)\right), \\
    R(w) &\leq R_S(w) + 96 \sqrt{\frac{m_2 h_1 LB}{n}} + C \delta^2 + \eta \quad \left(w \in B_{\delta}(w_2)\right), \\
    R(w) &\leq R_S(w) + 96 \sqrt{\frac{m_2 m_1 LB}{n}} + C \delta^2 + \eta \quad \left(w \in B_{\delta}(w_3)\right).
\end{align*}
Our bound adapts to the local geometry and is invariant under smooth reparameterizations unlike norm-based bounds.
Norm-based bounds \cite{Neyshabur2015-er, Golowich2018-sj, Ledent2025-fb}, a central theory for analyzing the generalization error of neural networks, depend on weight norms; however, it has been pointed out that these norms can change under function-preserving reparameterizations \cite{Dinh2017-wt}.
In contrast, our bound is invariant under (smooth) reparameterizations.
Indeed, since local RLCTs are invariant under diffeomorphisms (Proposition~\ref{lem:diffeomor_invarient}), the same bound holds if $\delta$ is re-chosen appropriately.
Because our bound is local, it is not directly comparable to these bounds; nevertheless, it depends only on the intrinsic local geometry.

\section{Experiment}\label{sec:experiments}

In this section, we empirically verify Main Theorem~\ref{thm:hessian_rank_ineq}.

\subsection{Verification of the Hessian Rank Bound in Simple Models}

We verify Main Theorem~\ref{thm:hessian_rank_ineq} experimentally 
for the concrete examples.
Specifically, for models for which the local RLCT is known as Section \ref{subsec:rrr} and \ref{subsec:lnn}, we numerically compare the rank of the Hessian with the local RLCT.
For the reduced-rank regression setup (Section~\ref{subsec:rrr}),
let $r = 1, h=2, m_1 = 3$ and $m_2 = 10$ in \eqref{eq:rrr-rlct}, then we have 
$\lambda(w_1)=7,  \lambda(w_2)=\frac{21}{2}, $ and $ \lambda(w_3)=7$.
For the linear neural network setup (Section~\ref{subsec:lnn}),
let $m_1 = 5, h_1=3, h_2=2$ and $m_2 = 1$ in \eqref{eq:lnn-rlct}, then 
$\lambda(w_1)=\frac{1}{2}, \lambda(w_2)=\frac32$ and  $\lambda(w_3)=\frac{5}{2}$.
In both experiments, we assume $\mathbb{E}[XX^{\top}]=I_{m_1}$, which satisfies the positive definiteness assumption on $\mathbb{E}[XX^{\top}]$ in Sections~\ref{subsec:rrr} and \ref{subsec:lnn}.
Then, the generalization error is given by $R(w) = \|UV - T^\star\|_F^2$ for reduced-rank regression and $R(w) = \|W_{\mathrm{out}} W_{\mathrm{hid}} W_{\mathrm{in}}\|_F^2$ for the linear neural network.
All eigenvalue computations were performed in double precision (float64).
Note that no samples need in this setting.

\begin{figure}
    \centering
    {\textbf{Reduced Rank Regression}}
    \includegraphics[width=\linewidth]{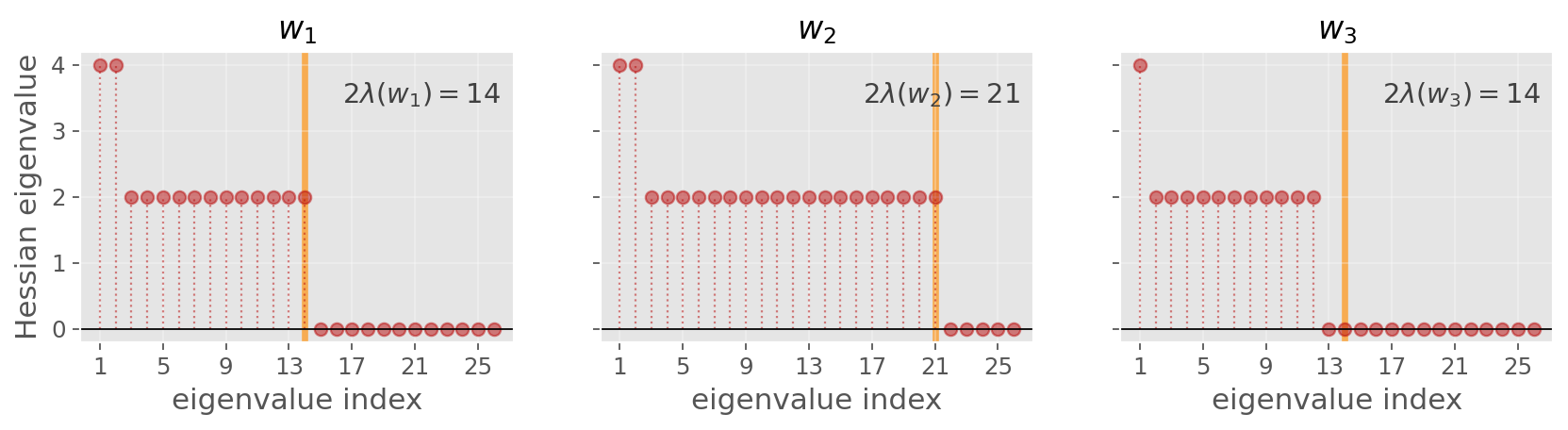}\par
    {\textbf{Linear Neural Network}}
    \includegraphics[width=\linewidth]{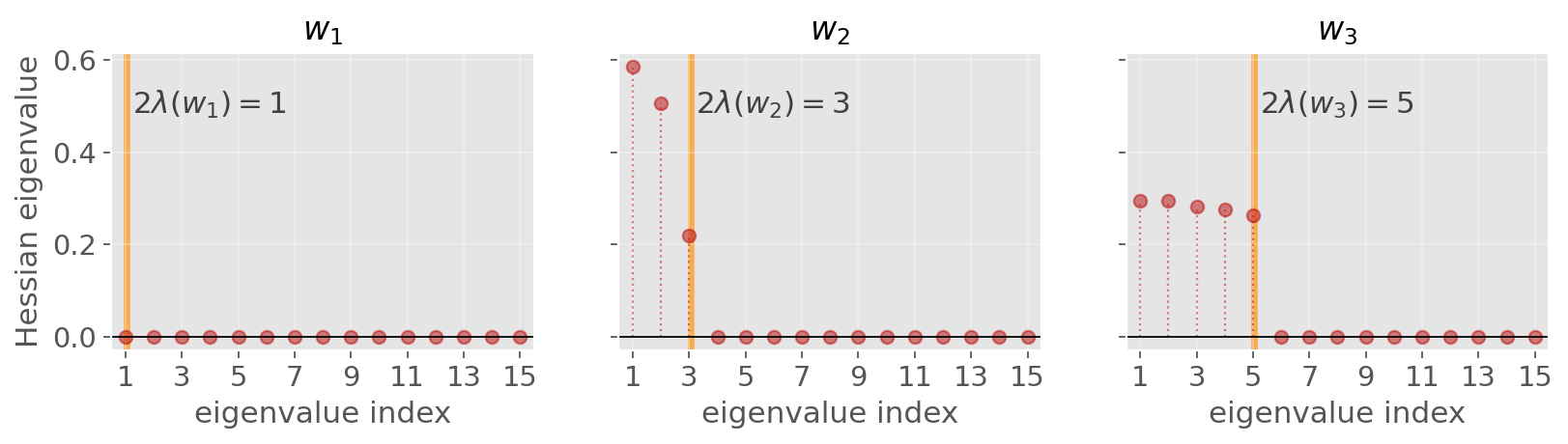}
    \caption{
        \textbf{Hessian Eigenvalues vs. Local RLCT}:
        $\operatorname{rank}\nabla^2 R(w_i) \leq 2\lambda(w_i) \; (i=1,2,3)$ holds at all displayed parameter values. 
        The red dots indicate the values of the eigenvalues, and the orange vertical line marks $2\lambda (w_i)$.
        The rank of the Hessian equals the number of its positive eigenvalues at global minimizers.
        The parameters and their corresponding local RLCTs are given in \eqref{eq:rrr-rlct} for reduced-rank regression, and in \eqref{eq:def_w_lnn} for the linear neural network.
    }
    \label{fig:RRR_hessian_rank_boumd}
\end{figure}

Fig.~\ref{fig:RRR_hessian_rank_boumd} shows that the rank of the Hessian is well bounded by $2\lambda(w_i)$ ($i=1,2,3$).
In reduced rank regression, the local RLCT at $w_3$ differs from the rank of the Hessian. This is because the local RLCT also captures contributions of order higher than two \cite{Wang2024-xw}, which the Hessian cannot detect.

\subsection{More General Networks}\label{sec:gen_net}

We conduct experiments for cases in which the local RLCT is difficult to compute theoretically.
The local RLCT can be estimated by the method proposed in \cite{lau2023local} whose code is available from the GitHub repository given in the paper.

As a basic setup, we consider a GELU or ReLU neural network $\phi_w:\mathbb{R}^3 \to \mathbb{R}$. 
The number of hidden layers and width  indicated in the graph.
We set the true function to $Y=0$, which allows us to identify parameter that realize it.

To study the local RLCT, we select the global minimizers as follows.
We construct five parameters $w_i, i=1,\ldots5$, each representing the zero function. 
Roughly speaking, we set the outputs of all units in the last layer to zero.
Specifically, each hidden unit has the form $w_{\mathrm{out}}\,\sigma(w_{\mathrm{in}}^{\top} x)$, where $\sigma$ is an activation function with $\sigma(0)=0$, $x$ is the input or the output of the previous layer, $w_{\mathrm{in}}$ is a vector of the same dimension as $x$, and $w_{\mathrm{out}}$ is a scalar.
If $w_{\mathrm{out}} = 0$, or $w_{\mathrm{in}} = 0$, then $w_{\mathrm{out}}\,\sigma(w_{\mathrm{in}}^{\top} x)=0$. 
Assume that the last hidden layer has width at least $5$.
For $i = 1, \ldots, 5$, the parameter $w_i$ is defined so that the first $j-1$ units of the last hidden layer have $w_{\mathrm{in}} = 0$ and the remaining units have $w_{\mathrm{out}} = 0$.
All remaining parameters are assumed to be drawn randomly from a Gaussian distribution.
Then, all of these parameters represent the zero function.

We need to estimate the the local RLCTs, since they are not theoretically clarified.  
To estimate the local RLCT, we use the estimation method based on \cite{lau2023local}. 
We draw $N=1024$ independent inputs $X_k\sim N(0,I_3)$ and use
$R_N(w)=\frac1N\sum_{k=1}^{N} \ell(X_k,w)$ as a Monte Carlo approximation of $R(w)$. 
Note that $R_N$ is used for estimating the generalization error, not for training.
At each $w_i$, we compute the eigenvalues of $\nabla^2R_N(w_i)$ and estimate the local RLCT using the WBIC-type estimator \cite{lau2023local}
\[
\widehat\lambda(w_i)
=
N \beta\left\{
\frac1T\sum_{t=1}^{T}R_N(w_i^{(t)})-R_N(w_i)
\right\},
\qquad \beta=\frac1{\log N},
\]
where $w_i^{(t)}$ denotes the $t$-th sample retained after burn-in from the SGLD chain initialized at $w_i$, and $T=1750$ is the number of retained samples. 
We use 2500 steps with step size $3\times10^{-4}$, Gaussian prior scale $\gamma=1$, and batch size $N$, discarding the first 750 steps.

We plot the Hessian eigenvalues in descending order and indicate \(2\widehat\lambda(w_i)\) on the eigenvalue-index axis. This comparison examines how the Hessian eigenvalues and the estimated local RLCT characterize neighborhoods of different parameters representing the same function.

\subsubsection{GELU Network}\label{sec:gelu_exp}

\begin{figure}
    \centering
    {\textbf{GELU Network} (Number of hidden layers: 1, Hidden layer widths: 5)}
    \includegraphics[width=\linewidth]{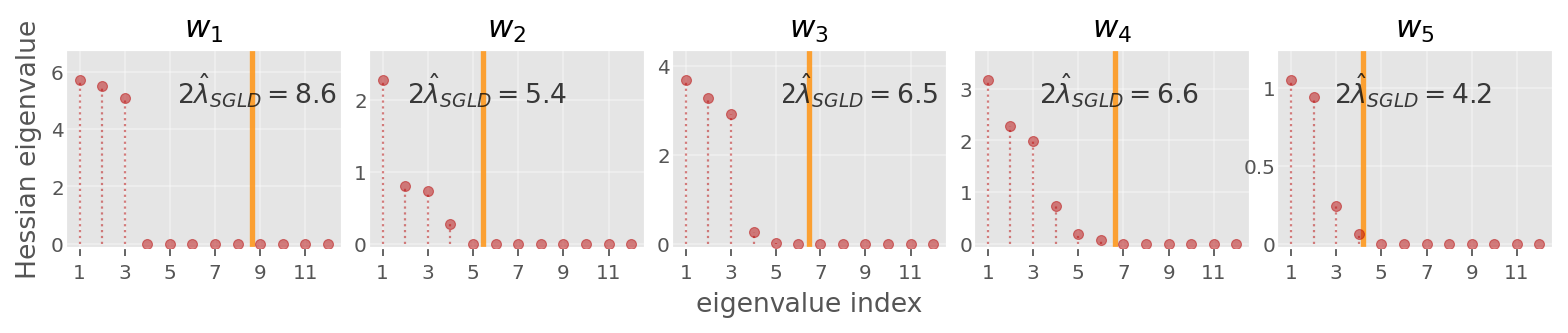}

    \centering
    {\textbf{GELU Network} (Number of hidden layers: 1, Hidden layer widths: 10)}
    \includegraphics[width=\linewidth]{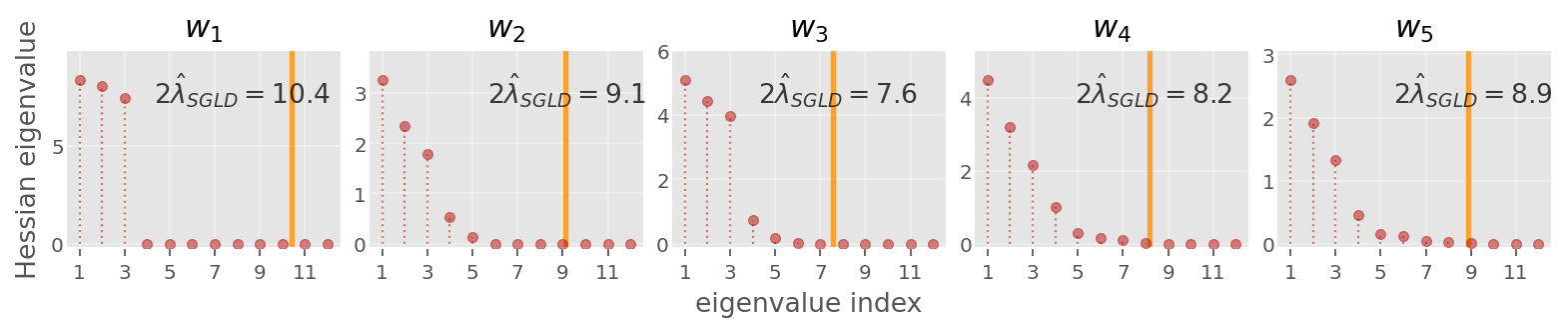}

    \centering
    {\textbf{GELU Network} (Number of hidden layers: 5, Hidden layer widths: (3,4,5,6,7))}
    \includegraphics[width=\linewidth]{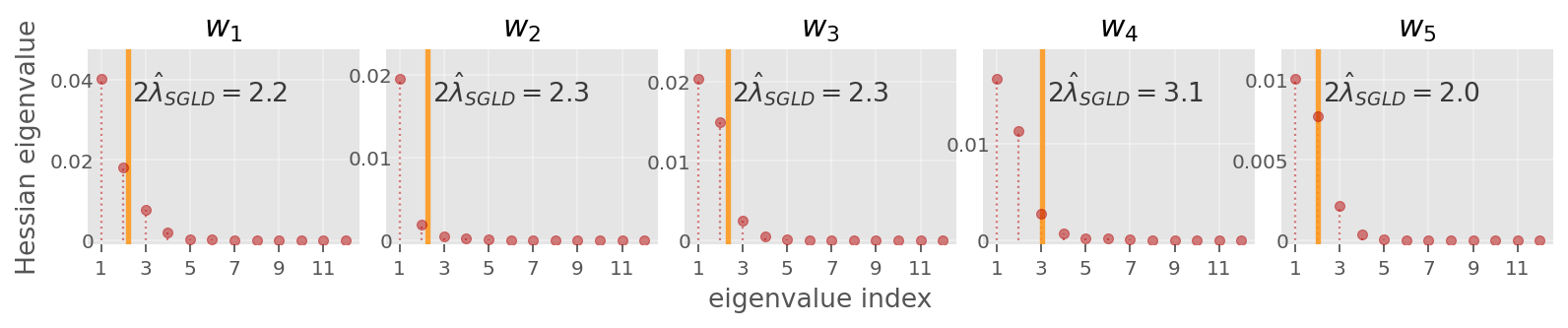}
    \caption{
        \textbf{Hessian Eigenvalues vs. Local RLCT (GELU Network)}:
            This plot shows the eigenvalues of the Hessian and the estimated local RLCTs $\hat{\lambda}_{\mathrm{SGLD}}(w_i)$ (orange lines) at five parameters $w_i \; (i=1,\ldots,5)$, which are chosen from a GELU network so that the output function is $0$.
            The number of dominant eigenvalues is roughly bounded above by $2 \lambda(w_i) \; (i=1,\ldots, 5)$.
    }
    \label{fig:GELU_hessian_rank_boumd}
\end{figure}

We conduct experiments using GELU networks (Fig.~\ref{fig:GELU_hessian_rank_boumd}). 
Despite some fluctuations, the results suggest that the local RLCT roughly upper-bounds the number of large eigenvalues.

The advantage of this method is that it can directly estimate an upper bound on the rank of the Hessian. 
As shown in the Fig.~\ref{fig:GELU_hessian_rank_boumd}, it is difficult to determine the exact rank of the Hessian because its eigenvalues decay gradually.
In fact, in numerical experiments, the Hessian often has many positive eigenvalues that are very close to zero \cite{Xie2022-xf, Karakida2021-ny}.
On the other hand, since $2 \lambda(w_i) \; (i=1,\ldots, 5)$ is guaranteed to be an upper bound on the rank of the Hessian, it can provide a criterion for deciding which
eigenvalues may be treated as zero and ignored.

\subsubsection{ReLU Network}

\begin{figure}
    \centering
    {\textbf{ReLU Network} (Number of hidden layers: 1, Hidden layer widths: 5)}
    \includegraphics[width=\linewidth]{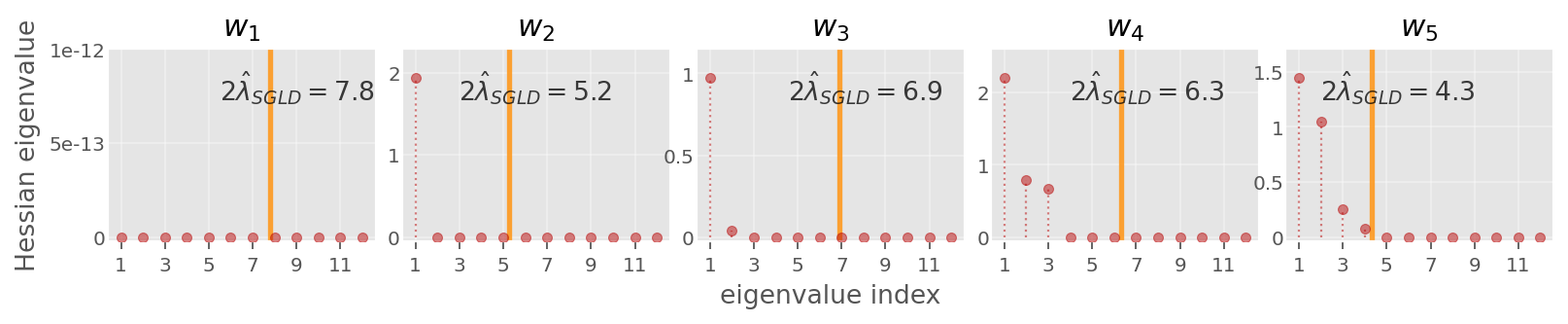}

    \centering
    {\textbf{ReLU Network} (Number of hidden layers: 1, Hidden layer widths: 10)}
    \includegraphics[width=\linewidth]{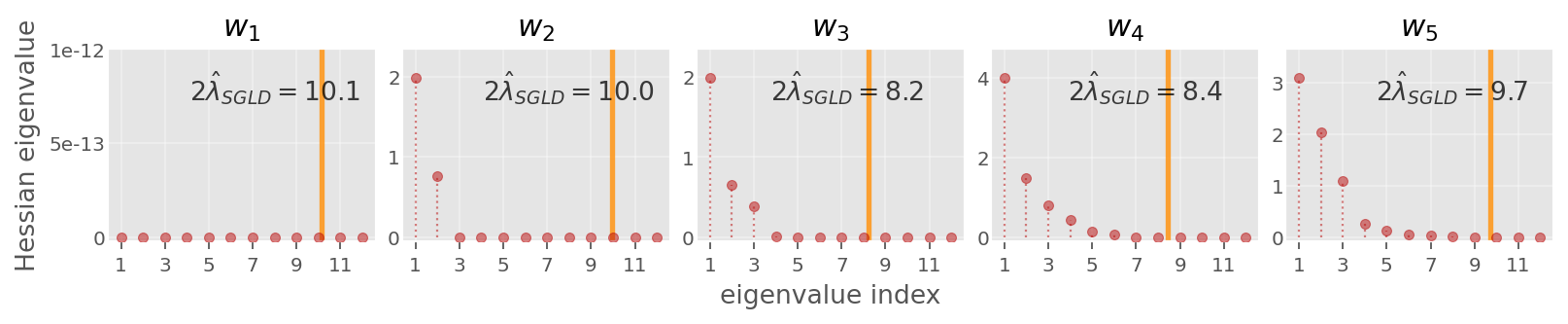}

    \centering
    {\textbf{ReLU Network} (Number of hidden layers: 5, Hidden layer widths: (3,4,5,6,7))}
    \includegraphics[width=\linewidth]{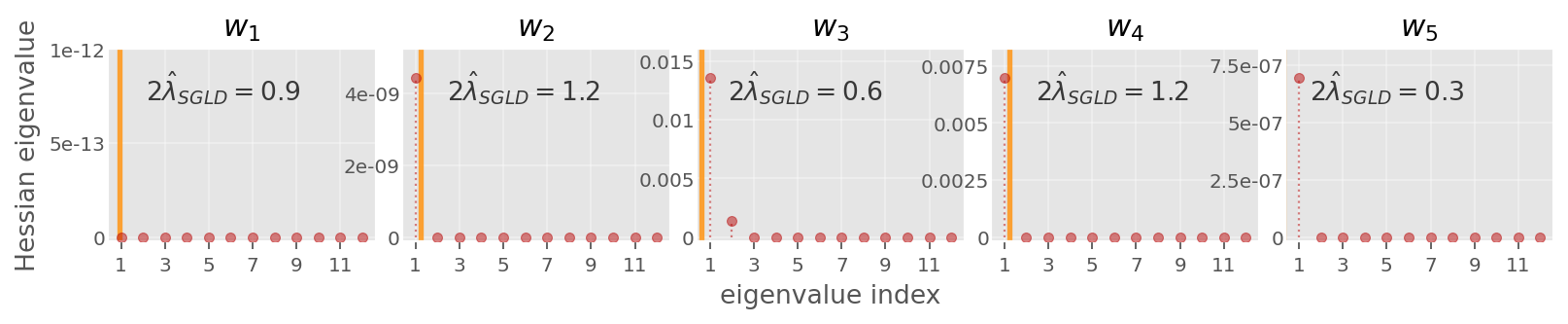}
    \caption{
        \textbf{Hessian Eigenvalues vs. Local RLCT (ReLU Network)}:
            This plot shows the eigenvalues of the Hessian and the estimated local RLCTs $\hat{\lambda}_{\mathrm{SGLD}}(w_i)$ (orange lines) at five parameters $w_i \; (i=1,\ldots,5)$, which are chosen from a ReLU network so that the output function is zero.
            Since ReLU is not analytic, the experimental results do not clearly support the claim of the Main Theorem.
    }
    \label{fig:ReLU_hessian_rank_boumd}
\end{figure}

We investigate what happens when the assumption~\ref{asmp:analytic} is not satisfied.
The loss function must be analytic for Main Theorem~\ref{thm:hessian_rank_ineq},
however, one of the typical activation functions ReLU is not analytic, since it is not smooth at the origin. 
The experimental results do not clearly support the claim (Fig.~\ref{fig:ReLU_hessian_rank_boumd}).

\section{Conclusion}
In this paper, we investigated the relationship between the local RLCT and the generalization error through local covering numbers.
We focused on the RLCT and its localized counterpart, the local RLCT, and showed that the latter can serve as a measure of local complexity even for models trained by empirical risk minimization, extending the role of RLCT beyond the characterization of Bayesian generalization behavior.

The main contribution of this work is twofold. 
(i) we proved that the local RLCT provides an upper bound on half the Hessian rank at local minimizers of the generalization error. 
(ii) we derived a local covering number bound whose exponent is governed by the local RLCT.
As an application of this result, 
we obtained a local uniform generalization gap bound. 
These results suggest that the local RLCT partially captures the effective dimension or intrinsic model complexity.

There remain several directions for future work. 
One important direction is to extend the analysis beyond neighborhoods of global minimizers of the generalization error and to develop analogous complexity bounds at any local minimizers. 
Another direction is to treat non-analytic models, especially ReLU networks, which are typical model in deep learning.

\bmhead{Acknowledgements}
Atsuyoshi Muta was supported by RIKEN Junior Research Associate Program. 
Masaaki Imaizumi was supported by JSPS KAKENHI (Grant No. 24K02904), JST CREST (Grant
No. JPMJCR21D2), JST FOREST (Grant No. JPMJFR216I), and JST BOOST (Grant
No. JPMJBY24A9).

\begin{appendices}

\section*{Additional Notation in the Appendix}

The notation and setup introduced in Sections~\ref{sec:notation} and~\ref{sec:basic_setup} are also used throughout this appendix.
The empirical measure $P_n$ is defined by
$P_n(A) = \frac{1}{n}\sum_{i=1}^n \mathbf{1}_{\{Z_i \in A\}}$
for all Borel measurable subset $A \subset \mathcal{Z}$ and $\mathbf{1}_{\{Z_i \in A\}}$ equals $1$ if $Z_i \in A$ and $0$ otherwise.

For $x \in \mathbb{R}^k$ and $\mathcal{S} \subset \mathbb{R}^k$, we define the distance between the point $x$ and the set $\mathcal{S}$ by $d(x,\mathcal{S}) = \inf_{y \in \mathcal{S}} \|x-y\|$.
For a differentiable function $g:\mathbb{R}^{k_1}\to\mathbb{R}^{k_2}  \; (k_1, k_2 \in \mathbb{N})$ , we denote by
$J_g (x)\in\mathbb{R}^{k_1 \times k_2}$ its Jacobian matrix at $x$, whose $(i,j)$-th
entry is given by
$(J_g(x))_{ij}=\frac{\partial g_i(x)}{\partial x_j}$.

Let $\mathbb{N}_0 := \{0\} \cup \mathbb{N}$. 
For $r \in \mathbb{N}$, an element $h = (h_1,\ldots,h_r) \in \mathbb{N}_0^r$ is called a multi-index of length $r$. 
For $x = (x_1,\ldots,x_r) \in \mathbb{R}^r$, we define $x^h = x_1^{h_1}\cdots x_r^{h_r}$. By a slight abuse of notation, we use the same notation for the gradient operator $\nabla = (\partial/\partial x_1,\ldots,\partial/\partial x_r)$, and set $\nabla^h = (\partial/\partial x_1)^{h_1}\cdots(\partial/\partial x_r)^{h_r}$. 
When it is necessary to specify the variable of differentiation, we write $\nabla_x$; otherwise, $\nabla$ denotes differentiation with respect to all variables. 
We define the order of $h \in \mathbb{N}_0^r$ by $\|h\|_1 := \sum_{j=1}^r |h_j| = h_1+\cdots+h_r$, i.e. the $L^1$-norm of $h:\{1,\dots,r\} \to \mathbb{N}_0$ for the counting measure.

\section{Analyticity of Generalization Error} \label{subsec:gen_err_analytic}
With Assumption \ref{asmp:analytic}, we obtain the following statement.
\begin{proposition}\label{prop:gen_err_analytic}
Under Assumption~\ref{asmp:analytic}, $R(w)$ is analytic on $W$.
\end{proposition}
\begin{proof}
We show that $R(w)$ is analytic.
Let $K = m+d$.
By Assumption~\ref{asmp:analytic}, for any $p=(\bar z, \bar w) \in \mathcal{O}_{\mathcal Z} \times W$, there exists $r>0$ such that $\ell$ admits the following power series expansion on $B_r^K(p)$:
\[
 \ell(z,w)
  =
 \sum_{\alpha \in \mathbb{N}_0^{K}}
 \frac{\nabla^{\alpha} \ell(\bar z,\bar w)}{\alpha!}
 h^{\alpha} \qquad (h \in B_r^K(p)).
\]
Here and below, $h=(z- \bar z,w- \bar w)$.  We write a multi-index $\alpha\in\mathbb N_0^K$ as $\alpha=(\gamma,\beta)$, where $\gamma\in\mathbb N_0^{m}$ corresponds to the $z$-variables and $\beta\in\mathbb N_0^d$ corresponds to the $w$-variables.

Fix $\bar w \in W$.  
Since $\mathcal Z\times\{\bar w\}$ is compact,  we may cover $\mathcal Z\times\{\bar w\}$ by finitely many balls on which the above power series expansions converge absolutely and uniformly on smaller balls.
Hence, there exist $z_1,\dots,z_N\in\mathcal Z$ and $\rho>0$ such that, for every $(z, w) \in \mathcal Z \times B_\rho^d(\bar w)$ there exists $i\in\{1,\dots,N\}$ for which $z$ belongs to the corresponding neighborhood of \(z_i\), and the associated power series expansion converges absolutely and uniformly.
\begin{align*}
\ell(z,w)
&=
\sum_{\gamma\in\mathbb N_0^{m}}
\sum_{\beta\in\mathbb N_0^d}
\frac{
\nabla_z^ \gamma \nabla_w^\beta \ell(z_i,\bar w)
}{\gamma!\beta!}
(z-z_i)^\gamma (w- \bar w)^\beta,\\
&=
\sum_{\beta\in\mathbb N_0^d}
\frac{1}{\beta!}
\left(
\sum_{\gamma\in\mathbb N_0^{m}}
\frac{
\nabla_z^\gamma \nabla_w^\beta \ell(z_i,\bar w)
}{
\gamma!
}
(z-z_i)^\gamma
\right)
(w-\bar w)^\beta
\\
&= 
\sum_{\beta\in\mathbb N_0^d} \frac{\nabla_w^\beta \ell(z, \bar w)}{\beta!} (w- \bar w)^\beta.
\end{align*}
Hence, for every \(w\in B_\rho^d(\bar w)\), we may interchange the summation and
the integral to get
\begin{align*}
 R(w)
 &= \int_{\mathcal Z}\ell(z,w) d P(z) \\
 &= \int_{\mathcal Z}
    \sum_{\beta\in\mathbb N_0^d}
    \frac{\nabla_w^{\beta}\ell(z,\bar w)}{\beta!} (w- \bar w)^\beta
     d P(z) \\
&= \int_{\mathcal Z}
    \sum_{i=0}^{\infty}
    \sum_{\|\beta\|_1=i}
    \frac{\nabla_w^\beta \ell(z,\bar w)}{\beta!}(w- \bar w)^\beta
     d P(z) \\
&= \lim_{J\to \infty}\int_{\mathcal Z}
    \sum_{i=0}^{J}
    \sum_{\|\beta\|_1=i}
    \frac{\nabla_w^\beta \ell(z,\bar w)}{\beta!}(w-\bar w)^\beta
     d P(z) \\
&= \sum_{\beta\in\mathbb N_0^d}
    \left(
      \int_{\mathcal Z}
      \frac{\nabla_w^\beta \ell(z,\bar w)}{\beta!} d P(z)
    \right)
    (w-\bar w)^\beta.
\end{align*}
The exchange of the limit and the integral is justified by the absolute uniform convergence of the power series and Lebesgue's dominated convergence theorem.
The right-hand side is a power series in $w- \bar w$ with positive radius of convergence.  
Therefore $R$ is real analytic in a neighborhood of $\bar w$.  
Since $\bar w \in W$ was arbitrary, $R$ is real analytic on $W$.
This proves the proposition.
\end{proof}

\section{Full Proof}\label{sec:full_proof}

\subsection{Proof of Lemma~\ref{lem:splitting}}\label{sec:prf_spliting}
\begin{proof}
Since $R(w) - R(w_*) \ge 0$ in an appropriate neighborhood of $w_*$, we have $\nabla R(w_*)=0$ and  $\nabla^2 R(w_*)$ is positive semidefinite.
Since $\operatorname{rank}\nabla^2 R(w_*)=r$, after a linear change of coordinates we may write
$w=(u,v)\in \mathbb{R}^r\times \mathbb{R}^{d-r}$,
so that
\[
    \nabla^2R(u_*, v_*)
    =
    \begin{pmatrix}
        \Lambda_+ & 0\\
        0 & 0
    \end{pmatrix},
\]
where $\Lambda_+$ is a symmetric positive definite $r\times r$ matrix.
Since
\[
    \nabla_u R(u_*,v_*)=0,
    \qquad
    \nabla_u^2 R(u_*,v_*)=\Lambda_+,
\]
and $\Lambda_+$ is invertible, the implicit function theorem implies that there exists a $C^2$ map $\sigma(v)$ defined around $v_*\in\mathbb{R}^{d-r}$ such that $\sigma(v_*)=u_*$ and
\[
    \nabla_u R(\sigma(v),v)=0.
\]

Define $\widetilde R(u,v) = R(u+\sigma(v),v)$.
Then $\nabla_u \widetilde R(0,v) = \nabla_u R(\sigma(v),v) =0$.
Therefore, by Taylor's formula in the $u$-variables around $u=0$,
\[
    \widetilde R(u,v)
    =
    \widetilde R(0,v)
    +
    \left\langle B(u,v) u,u\right\rangle,
\]
where
\[
    B(u,v)
    :=
    \int_0^1 (1-t) 
    \nabla_u^2 \widetilde R(tu,v) dt .
\]
The matrix $B(u,v)$ is symmetric and continuous, and
$B(0,v_0)=\frac12 \Lambda_+$.
Hence, $B(u,v)$ remains symmetric positive definite in a sufficiently small neighborhood of the $(0, v_*)$.
Let
$C(u,v)=B(u,v)^{1/2}$, and $\bar u = C(u,v)u$,
then we have
\[
    \widetilde R(u,v)
    =
    \widetilde R(0,v) + \|\bar u\|^2
\]
around $(0,v_*)$.
Define $\mathrm{Res}(v)=\widetilde R(0,v)=R(\sigma(v),v)$.
Renaming $\bar u$ as $u$, we get
\[
    R\circ \Phi(u,v)
    =
    \|u\|^2+\mathrm{Res}(v),
\]
for some $C^1$ diffeomorphism $\Phi$ around $w_*$.

It remains to check ${\mathrm{Res}}(v_*) = R(w_*), \nabla \mathrm{Res}(v_*)=0$ and $\nabla^2 {\mathrm{Res}}(v_*)=0$.
Since $\sigma(v_*)=u_*$, we have $\mathrm{Res}(v_*)=R(u_*,v_*)=R(w_*)$.
Moreover, differentiating $\nabla_u R(\sigma(v),v)=0$ at $v_0$ gives
\[\Lambda_+ \nabla_v \sigma(v_*)+\nabla_v\nabla_u R(u_*,v_*)=0.\]
Since $\nabla_v\nabla_u R(u_*,v_*)=0$ and $\Lambda_+$ is invertible,
$\nabla_v\sigma(v_*)=0$.
Therefore, noting that $\nabla R(w_*)=0$ implies $\nabla_u R(u_*,v_*)=0$ and $\nabla_v R(u_*,v_*)=0$,
\begin{align*}
    \nabla \mathrm{Res}(v_*)
    &=
    \nabla_u R(u_*,v_*) \nabla_v\sigma(v_*)
    +\nabla_v R(u_*,v_*)
    =0,\\
    \nabla^2\mathrm{Res}(v_*)
    &=
    \nabla_v\sigma(v_*)^\top
    \nabla_u^2R(u_*,v_*)
    \nabla_v\sigma(v_*)
    +\nabla_v^2R(u_*,v_*)
    =0,
\end{align*}

Finally, since $\mathrm{Res}(v)=R(\sigma(v),v)$
and $R \ge R(w_*)$ near $w_0$, we have $\mathrm{Res}(v)\ge R(w_*)$. 
This completes the proof.
\end{proof}

\subsection{Statement and Proof of Lemma~\ref{lem:loss_difference_bound}}\label{sec:loss_difference_bound}

Before stating the lemma, we introduce some notations.
Fix an arbitrary $w_0 \in W_0$, and let $r=\operatorname{rank} \nabla^2 R(w_0)$.
There exist matrices
\(Q_+\in\mathbb R^{d\times r}\) and \(Q_0\in\mathbb R^{d\times(d-r)}\) such that the matrix $Q=(Q_+ \;  Q_0)$ diagonalizes $\nabla^2 R(w_0)$.
That is, $Q$ is an orthogonal matrix, and 
\[
    Q^\top \nabla^2 R(w_0) Q
    =
    \begin{pmatrix}
        \Lambda_+ & 0\\
        0 & 0
    \end{pmatrix},
\]
where 
$\Lambda_+ =
    \operatorname{diag}
    \bigl(
        \lambda_1^+(w_0),\ldots,\lambda_r^+(w_0)
    \bigr)$ 
is the diagonal matrix whose diagonal entries are the positive eigenvalues of $\nabla^2 R(w_0)$.

For $u\in\mathbb R^r$ and $v\in\mathbb R^{d-r}$, define the affine
orthogonal coordinate map
\begin{align}\label{eq:linear_tran}
    \Psi(u,v)
    =
    w_0
    +
    Q_+ u
    +
    Q_0 v.
\end{align}
Then \(\Psi(0, 0)=w_0\).

\begin{lemma}[Loss Difference Bound]\label{lem:loss_difference_bound}
For any $w_0 \in W_0$, let $\Psi$ denote the coordinate transformation defined in \eqref{eq:linear_tran}.
Then there exist constants $L, C > 0$ and $\delta > 0$ such that for any $(u_1,v_1), (u_2,v_2) \in B_{\delta}((0, 0))$
\[
    \bigl\|
        \ell(\cdot, \Psi(u_1, v_1))-\ell(\cdot,\Psi(u_2, v_2))
    \bigr\|_{\infty}
    \le
    L \|u_1-u_2\|
    +
    C \delta\|v_1-v_2\|.
\]
In particular, noting that $\|v_1 -v_2\| \le 2\delta$,
\[
    \bigl\|
        \ell(\cdot, \Psi(u_1, v_1))
        -
        \ell(\cdot, \Psi(u_2, v_2))
    \bigr\|_{\infty}
    \le
    L \|u_1-u_2\|
    +
    2 C \delta^2.
\]
\end{lemma}

\begin{proof}
The main idea of the proof is to use the coordinate transformation $\Psi$ to obtain direction wise bounds.
By the fundamental theorem of calculus, we have
\begin{align*}
&\ell(\cdot, \Psi(u_1,v_1)) - \ell(\cdot, \Psi(u_2,v_2)) 
=\int_0^1 \frac{d}{dt} \ell(z, \Psi(u_t, v_t))   dt\\
&\qquad= \int_0^1 
    \left\{ 
    \langle \nabla_u \ell(z, \Psi(u_t, v_t)), u_1 - u_2 \rangle 
    + \langle \nabla_v \ell(z, \Psi(u_t, v_t)), v_1 - v_2 \rangle
    \right\}
  dt,
\end{align*}
where $(u_t, v_t) = t(u_1,v_1) + (1-t)(u_2,v_2)$ $(0\le t\le 1)$.
Then we show that there exist suitable constants $L, C > 0$ and $\delta > 0$ such that the following bound holds:
\begin{align*}
&
\bigl\|
    \ell(\cdot, \Psi(u_1,v_1))
    -
    \ell(\cdot, \Psi(u_2,v_2))
\bigr\|_{\infty}
\\
&\le
\int_0^1
\|\nabla_u \ell(\cdot,\Psi(u_t, v_t))\|_{\infty}
\|u_1-u_2\| dt
+
\int_0^1
\|\nabla_v \ell(\cdot, \Psi(u_t, v_t))\|_{\infty}
\|v_1-v_2\| dt
\\
&\le
L \|u_1-u_2\| + C \delta \|v_1-v_2\|.
\end{align*}
To this end, it remains to prove the following two inequalities.
\begin{enumerate}[label=(\roman*)]
    \item $\|\nabla_u \ell(\cdot,\Psi(u_t, v_t))\|_{\infty} \le L$ \, $(0\le t\le 1)$,
    \item $\|\nabla_v \ell(\cdot, \Psi(u_t, v_t))\|_{\infty} \le C \delta$  \, $(0\le t\le 1)$.
\end{enumerate}

For (i), noting that $\nabla_u \ell(z,\Psi(u, v)) = Q_+^{\top} \nabla_w \ell(z, w)$, we bound $\|Q_+^{\top} \nabla_w \ell(z, w)\|$.
By Taylor's theorem, there exists $\delta_1 > 0$ such that for all $w \in B_{\delta_1}(w_0)$,
\begin{align*}
\left\|Q_+^{\top} \nabla_w \ell(z, w) \right\|
&\le \left\|Q_+^{\top} \nabla_w \ell(z, w_0)\right\| 
+ \int_0^1 \left\| Q_+^{\top} \nabla_w^2 \ell(z, w_0 + t(w-w_0)) (w-w_0)\right\| \, dt, \\
&\le \left\|Q_+^{\top} \nabla_w \ell(z, w_0)\right\| 
+ \sup_{w \in B_{\delta_1}(w_0)}\left\| Q_+^{\top} \nabla_w^2 \ell(z, w)\right\| \delta_1\\
&\le \left\|Q_+^{\top} \nabla_w \ell(z, w_0)\right\| + 1.   
\end{align*}
Thus we define $L = \left\|Q_+^{\top} \nabla_w \ell(\cdot, w_0)\right\|_{\infty} + 1$.
This yields the desired conclusion.

For (ii), noting that $\nabla_v \ell(z, \Psi(u, v)) = Q_0^{\top} \nabla_w \ell(z, w)$, we bound $\|Q_0^{\top} \nabla_w \ell(z, w)\|$.
By Taylor's theorem, we have
\begin{align}\label{eq:taylor}
    Q_0^{\top} \nabla_w \ell(z, w) 
    &= Q_0^{\top} \nabla_w \ell(z, w_0) + Q_0^{\top} \nabla_w^2 \ell(z, w_0) (w-w_0)\notag\\ 
    &\qquad + Q_0^{\top} \sum_{|\beta|=2} \frac{1}{\beta!} \nabla_w^{\beta} (\nabla_w \ell)(z, w^\circ) (w-w_0)^{\beta},
\end{align}
where $w^\circ = w - \theta (w - w_0)$ for some $\theta \in (0,1)$.

Then for $v \in \mathrm{ker} \nabla^2R(w_0)$, we have
\begin{align*}
    \mathbb{E}_{Z \sim P} \bigl[\|v^{\top} \nabla_w \ell(Z,w_0)\|^2]
    &= \mathbb{E}_{Z \sim P} \bigl[\langle \nabla_w \ell(Z,w_0) \nabla_w \ell(Z,w_0)^{\top} v, v \rangle],\\
    &= \langle \mathbb{E}_{Z \sim P} \bigl[ \nabla_w \ell(Z,w_0) \nabla_w \ell(Z,w_0)^{\top}] v, v \rangle,\\
    &= \langle \mathrm{Cov}(\nabla_w \ell(Z,w_0)) v, v \rangle 
    \le \alpha \langle \nabla^2 R(w_0) v, v \rangle 
    = 0,
\end{align*}
where the inequality follows from Assumption~\ref{asmp:fim}.
Therefore, $v^{\top}\nabla_w \ell(Z,w_0)=0$ a.s., and by continuity of the loss function, $v^{\top}\nabla_w \ell(z,w_0)=0$ for every $z \in \mathcal Z$.
Since the columns of $Q_0$ form an orthonormal basis of $\mathrm{ker} \nabla^2 R(w_0)$, we obtain
$Q_0^{\top}\nabla_w \ell(z,w_0)=0$ for every $z \in \mathcal Z$.
Consequently, by \eqref{eq:taylor}, we obtain
\begin{align*}
    Q_0^{\top} \nabla_w \ell(z, w)
    &=  Q_0^{\top} \nabla_w^2 \ell(z, w_0) (w-w_0)  + Q_0^{\top} \sum_{|\beta|=2} \frac{1}{\beta!} \nabla_w^{\beta} (\nabla_w \ell)(z, w^\circ) (w-w_0)^{\beta},
\end{align*}
Then, there exist $\delta_2>0$ such that for all $w \in B_{\delta_2}(w_0)$
\begin{align*}
    \left\|Q_0^{\top} \nabla_w \ell(\cdot, w) \right\|
    &\le \left\| Q_0^{\top} \nabla_w^2 \ell(z, w_0) \right\| \|w-w_0\| + \left\|Q_0^{\top} \sum_{|\beta|=2} \frac{1}{\beta!} \nabla_w^{\beta} (\nabla_w \ell)(z, w^\circ)\right\| \|w-w_0\|^2 \\
    &\le \left( \left\| Q_0^{\top} \nabla_w^2 \ell(z, w_0) \right\|  + \left\|Q_0^{\top} \sum_{|\beta|=2} \frac{1}{\beta!} \nabla_w^{\beta} (\nabla_w \ell)(z, w^\circ)\right\| \|w-w_0\| \right) \|w-w_0\| \\
    &\le \left( \left\| Q_0^{\top} \nabla_w^2 \ell(z, w_0) \right\|  + 1 \right) \delta_2 \\
    &\le C\delta_2,
\end{align*}
where $C = 2 \left\| Q_0^{\top} \nabla_w^2 \ell(\cdot, w_0) \right\|_{\infty}  + 1$.
This yields the desired conclusion.

Finally, setting $\delta = \min\{\delta_1, \delta_2\}$ gives the desired conclusion.
\end{proof}

\subsection{Proof of Corollary~\ref{cor:local_gen_bound}}\label{sec:local_gen_bound}
\begin{proof}
We use the truncated form of truncated Dudley integral (Proposition~\ref{prop:dudley}).

We first define an upper bound $B$ on the loss function.
By taking $\delta$ small if necessary, we may assume that $\overline{B_\delta}(w_0) \subset W$.
Then, since $\overline{B_\delta}(w_0)$ is compact, we can define $B := \max_{(z,w) \in \mathcal{Z} \times \overline{B_\delta}(w_0)} \ell(z, w)$.
That is,  $\ell(z, w) \le B$ holds for all $(z, w) \in \mathcal{Z} \times B_\delta(w_0)$.

For every
$\alpha>0$ with $\alpha < B$,
\[
    \operatorname{Rad}(B_{\delta}(w_0))
    \le
    4\alpha
    +
    \frac{12}{\sqrt n}
    \int_\alpha^{2B}
    \sqrt{
        \log N\bigl(\mathcal{F}_{\ell}(B_{\delta}(w_0)),L^2(P_n),\tau\bigr)
    } d\tau.
\]

Let $r = \operatorname{rank} \nabla_w^2R(w_0)$, and first consider the case $r>0$.
By Main Theorem~\ref{thm:local_covering_bound},  
there exists a constant $\rho_\delta>0$ such that, for every $\tau>\rho_\delta$,
\begin{align*}
    N\bigl(\mathcal{F}_{\ell}(B_{\delta}(w_0)),L^2(P_n),\tau\bigr)
    &\le N\bigl(\mathcal{F}_{\ell}(B_{\delta}(w_0)),L^{\infty}(P),\tau\bigr)\\
    &\le
    \left(
        1+\frac{2L}{\tau-\rho_\delta}
    \right)^{2\lambda(w_0)} .
\end{align*}

Set $\alpha:=2\rho_\delta$.
Then \(\tau>\rho_\delta\) for all \(\tau\in[\alpha,2B]\), so the covering
bound above is applicable on the interval of integration. Therefore,
\[
\begin{aligned}
\int_\alpha^{2B}
    \sqrt{
        \log N\bigl(\mathcal{F}_{\ell}(B_{\delta}(w_0)),L^2(P_n),\tau\bigr)
    } d\tau  
\le
\sqrt{2\lambda(w_0)}
\int_\alpha^{2B}
    \sqrt{
        \log\left(
            1+\frac{2L}{\tau-\rho_\delta}
        \right)
    } d\tau .
\end{aligned}
\]
By the change of variables \(s=\tau-\rho_\delta\), we get
\[
\begin{aligned}
\sqrt{2\lambda(w_0)}
\int_\alpha^{2B}
    \sqrt{
        \log\left(
            1+\frac{2L}{\tau-\rho_\delta}
        \right)
    } d\tau 
\le
\sqrt{2\lambda(w_0)}
\int_0^{2B}
    \sqrt{
        \log\left(
            1+\frac{2L}{s}
        \right)
    } ds .
\end{aligned}
\]
Using the elementary inequality
$
\int_0^a
\sqrt{\log\left(1+\frac{c}{t}\right)} dt
\le
2\sqrt{ca}
$
 for $a>0,\ c>0$,
 we obtain
\[
\begin{aligned}
\int_\alpha^{2B}
    \sqrt{
        \log N\bigl(\mathcal{F}_{\ell}(B_{\delta}(w_0)),L^2(P_n),\tau\bigr)
    } d\tau
\le
4\sqrt{2\lambda(w_0)LB}.
\end{aligned}
\]
Thus the truncated Dudley bound gives
\[
\operatorname{Rad}(B_{\delta}(w_0))
\le
4 \alpha
+
48 \sqrt{\frac{2\lambda(w_0)LB}{n}}.
\]
Since \(\alpha=2\rho_\delta = 2 C \delta^2\), we have
\[
\operatorname{Rad}(A)
\le
8 C\delta^2
+
48
\sqrt{
    \frac{2\lambda(w_0)LB}{n}
}.
\]
By Proposition~\ref{prop:syutyu_ineq}, for every \(\eta>0\), with probability at least $1-\exp\left(-\frac{n\eta^2}{2B^2}\right)$,
we have
\[
\sup_{w\in B_{\delta}(w_0)}|R(w)-R_S(w)|
\le
2\operatorname{Rad}(B_{\delta}(w_0))+\eta .
\]
Therefore,
\[
\sup_{w\in B_{\delta}(w_0)}|R(w)-R_S(w)|
\le
16C\delta^2
+
96
\sqrt{
    \frac{2\lambda(w_0)LB}{n}
}
+\eta .
\]
Finally, by writing $16 C$ as $C$, we obtain
\[
\sup_{w\in B_{\delta}(w_0)}
|R(w)-R_S(w)|
\le
96
\sqrt{
    \frac{2\lambda(w_0)LB}{n}
}
+
C \delta^2
+
\eta .
\]

If \(r=0\), then the covering bound in Main Theorem~\ref{thm:local_covering_bound} is interpreted with
the right-hand side equal to \(1\). Hence the entropy integral over
\([\alpha,2B]\) is zero, and the same argument gives
\[
\sup_{w\in B_\delta(w_0)}
|R(w)-R_S(w)|
\le
C\delta^2+\eta ,
\]
which is stronger than the desired bound. This completes the proof.
\end{proof}

\section{On the Validity of Assumption~\ref{asmp:analytic}} \label{subsec:asmp_validity}

We provide a justification for Assumption~\ref{asmp:analytic}.
In the following two examples, we show that the loss function of representative tasks satisfies Assumption~\ref{asmp:analytic} when the learning model is analytic.
Here, a learning model $f(x, w)$ is said to be analytic if there exists an open set $\mathcal{O}_{\mathcal{X}}$ containing $\mathcal X$ and $f$ is analytic on $\mathcal{O}_{\mathcal{X}} \times W$.

\begin{example}[Regression]
In regression problems,
we verify that the squared loss regression satisfies Assumption~\ref{asmp:analytic}.
The model outputs 
\begin{align}\label{eq:regression}
f(x,w) = \bigl(f_1(x,w),\ldots,f_{m_2}(x,w)\bigr) \in \mathbb{R}^{m_2},    
\end{align}
are analytic on $\mathcal{O}_{\mathcal X} \times W$.
The loss function is defined by
\[
\ell((x,y),w)
=
\frac{1}{2}\|f(x,w)-y\|^2.
\]
$R$ is clearly analytic on $(\mathcal{O}_{\mathcal X} \times \mathcal{O}_{\mathcal Y}) \times W$ for any open set $\mathcal{O}_{\mathcal{Y}}$ containing $\mathcal Y$.
\end{example}

\begin{example}[Classification]
In classification problems, we define the output space $\mathcal{Y} = \{p \in [0,1]^{m_2} : \sum_{i=1}^{m_2} p_i = 1\}$.
The observed labels are one-hot vectors in $\{e_1, \ldots, e_K\} \subset \mathcal Y$, where $e_i$ $(i=1,\ldots,K)$ denotes the vector whose $i$-th component is $1$ and all other components are $0$.
We can then choose the open set $\mathcal{O}_{\mathcal{Y}}$ containing $\mathcal{Y}$ by $\mathcal O_{\mathcal Y} = (-\varepsilon, 1+\varepsilon)^{m_2}$ for sufficiently small $\varepsilon >  0$.
Consider a classification model whose outputs are
\begin{align}\label{eq:logit}
    f(x,w)=(f_1(x,w),\dots,f_{m_2}(x,w))\in \mathcal Y,
\end{align}
where $f_i(x,w)$ $(i=1,\ldots,m_2)$ are analytic on $\mathcal{O}_{\mathcal X} \times W$.
Then the cross-entropy loss can be written as 
\begin{align*}
    \ell((x,y),w)
    &=
    - \sum_{i=1}^{m_2} y_i \log f_i(x,w),
\end{align*}
and this is analytic on $(\mathcal{O}_{\mathcal X} \times \mathcal{O}_{\mathcal Y}) \times W$.
This shows Assumption~\ref{asmp:analytic} is satisfied on $(\mathcal{O}_{\mathcal X} \times \mathcal{O}_{\mathcal Y}) \times W$. 

\end{example}

Many models are analytic with respect to both the parameters and the samples. 
For example, in neural networks, classical activation functions such as \(\tanh\) and the sigmoid function are analytic.
Although the widely used ReLU is not analytic, GELU, which was proposed as an analytic alternative, has become widely used \cite{radford2018improving, devlin2019bert}.

\section{Regarding Assumption~\ref{asmp:fim}} \label{sec:reg_asmp1}

In addition to the negative log-likelihood loss, squared loss regression with bounded noise is another example that satisfies Assumption~\ref{asmp:fim}.

\begin{example}[Squared loss regression with bounded outputs]
Let the output space be $\mathcal Y=[-M,M]$ for some $M>0$.
Let $f_{w^\star}: \mathcal X \to \mathcal Y$ be the true function, and
\[
Y=f_{w^\star}(X)+\varepsilon,
\qquad
\mathbb E[\varepsilon\mid X]=0,
\]
where the noise variable $\varepsilon$ satisfies
$f_{w^\star}(X)+\varepsilon \in [-M,M]$.
We consider the squared loss
\[
\ell((x,y),w)
=
\frac12\bigl(y-f_w(x)\bigr)^2.
\]
\end{example}

\begin{proof}[Proof that this example satisfies Assumption~\ref{asmp:fim}.]
By the assumption on the noise, we have $|\varepsilon| = |Y - f_{w^\star}(X)| \le 2M$.
Noting that $\mathbb{E}[\nabla_w \ell(Z, w_0)] = \mathbb{E}[- \varepsilon \nabla_w f_{w_0}(X)] = 0$ holds for every $w_0 \in W_0$, we have
\begin{align*}
\nabla^2R(w_0)
&=
\mathbb E\left[
    \nabla_w f_{w_0}(X)
    \nabla_w f_{w_0}(X)^\top
\right]
-
\mathbb E\left[
    \varepsilon\nabla_w^2f_{w_0}(X)
\right] \\
&=
\mathbb E\left[
    \nabla_w f_{w_0}(X)
    \nabla_w f_{w_0}(X)^\top
\right],
\end{align*}
and, 
\begin{align*}
\mathrm{Cov}\left(\nabla_w\ell(Z,w_0)\right)
&=
\mathbb E\left[
    \nabla_w\ell(Z,w_0)
    \nabla_w\ell(Z,w_0)^\top
\right] \\
&=
\mathbb E\left[
    \varepsilon^2
    \nabla_w f_{w_0}(X)
    \nabla_w f_{w_0}(X)^\top
\right] \\
&\preceq
4M^2
\mathbb E\left[
    \nabla_w f_{w_0}(X)
    \nabla_w f_{w_0}(X)^\top
\right].
\end{align*}
Therefore, we have
\[
\mathrm{Cov}\left(\nabla_w\ell(Z,w_0)\right) \preceq 4M^2 \nabla^2R(w_0).
\]
Thus Assumption~\ref{asmp:fim} holds with $\alpha=4M^2$.
\end{proof}

\section{Basic Properties in Singular Learning Theory}\label{appendix:slt}

\subsection{Existence of the Local RLCT}

In this section, we establish the existence of the limit required to define the local RLCT.

\begin{lemma} \label{lem:def_of_LRLCT}
For any local minimizer $w_* \in W$ of $R$, there exists $\delta > 0$ such that
\[
\lambda\bigl(R, \overline{B_{\delta}}(w_*)\bigr) = \lambda\bigl(R, \overline{B_{\delta'}}(w_*)\bigr) \quad \text{for all } 0 < \delta' \le \delta.
\]
\end{lemma}

\begin{proof}
Let $f(w) = (R(w) - R(w_*))^2$, then by Hironaka's resolution of singularities \cite{hironaka1964resolution, watanabe2009algebraic}, there exist a real analytic manifold $\mathfrak{M}$ and a real analytic map $g:\mathfrak{M} \to W$ satisfying the following conditions:
Denoting $\mathfrak{M}_0 = \{u \in \mathfrak{M} : f \circ g(u) = 0\}$,
then for any $u \in \mathfrak{M}_0$,
\begin{align*}
    f \circ g (u) &= u^{2k_{\alpha}},\\
    |J_g(u)| &= b_{\alpha}(u) u^{h_{\alpha}},
\end{align*}
where $k_{\alpha}, h_{\alpha} \in \mathbb{N}_0^d$ and $b_{\alpha}(u)$ is a positive real analytic function.
In this case, using a coordinate neighborhood system $\{U_{\alpha}\}$ on $\mathcal M$, the $\zeta$ function on $C$ satisfies
\begin{align*}
    \zeta(s) 
    &= \int_{\overline{B_{\delta}}(w_*)} (R(w) - R(w_*))^{2s} dw, \\
    &= \int_{g^{-1} \left( \overline{B_{\delta}}(w_*) \right)} f \circ g(u)^{s} |J_g(u)| du, \\
    &= \sum_{\alpha} \int_{U_{\alpha} \cap g^{-1} \left(\overline{B_{\delta}} \left(w_* \right) \right)} u^{2 k_{\alpha} s + h_{\alpha}} b_{\alpha}(u) du.
\end{align*}
Therefore, the only candidates for the poles of $\zeta$ are $\left\{ - \frac{h_{\alpha_i} + 1}{2 k_{\alpha_i}} : i = 1, \cdots, d \right\}$.
By the inclusion of the integration domains, if $\delta' < \delta$ then $\lambda \left(R, \overline{B_{\delta}}(w_*) \right) \le \lambda \left(R,  \overline{B_{\delta'}}(w_*)\right)$, 
thus $\lambda(\overline{B_{\delta}}(w_*))$ is nonincreasing in $\delta$.
Moreover, since the candidates of RLCT $\left\{ \frac{h_{\alpha i} + 1}{2 k_{\alpha i}} : i = 1, \cdots, d \right\}$ is finite, we obtain the claim.
\end{proof}

\subsection{Some Important Properties of RLCT and Local RLCT}

In this subsection, we state some important properties of the RLCT, which also hold for the local RLCT.

First, the RLCT satisfies a monotonicity property with respect to functions.
This fact plays a central role in the proof of the main result.

\begin{proposition}[Monotonicity] \label{lm:rlct}
	Let $f, g: O \to \mathbb{R}$ be nonnegative real analytic functions defined on an open set $O \subset \mathbb{R}^d$.
	If $f \leq g$, that is, $f(x) \leq g(x)$ holds at every point $x \in O$, then for any compact set $C \subset O$,
    \[
        \lambda(f, C) \leq \lambda(g, C).
    \]
\end{proposition}

\begin{proof}
	Let $s$ be a negative real number. Then
	\[\int_C g(w)^{s} dw \leq \int_C f(w)^{s} dw,\]
	and hence, if the right-hand side converges, then the left-hand side must also converge.
	By the definition of the real log canonical threshold, the right-hand side converges for any $s$ such that $s > - \lambda(f, C)$.
	Therefore, we must have $- \lambda(g, C) \leq - \lambda(f, C)$.
	That is, we obtain $\lambda(f, C) \leq \lambda(g, C)$.
\end{proof}

The following proposition is useful for computing local RLCTs in practice.

\begin{proposition}[Invariance under Multiplicative Equivalence]\label{prop:muleq}
Let $f, g: O \to \mathbb{R}$ be nonnegative real analytic functions defined on an open set $O \subset \mathbb{R}^d$.
If there exist constans $c, c' > 0 $ such that $cg \le f \le c'g$, then 
\[
    \lambda(f, C) = \lambda(g, C).
\]
\end{proposition}

\begin{proof}
This follows from Proposition~\ref{lm:rlct} and the invariance of the convergence of the $\zeta$ function under multiplication by a positive constant.
\end{proof}

Moreover, the RLCT is invariant under diffeomorphic transformations of the parameter space.

\begin{proposition}[Diffeomorphism invariance]\label{lem:diffeomor_invarient}
    The RLCT is invariant under diffeomorphisms.
    That is, for a nonnegative real analytic function $f: O \to \mathbb{R}$ defined on an open set $O \subset \mathbb{R}^d$,
	when there exist an open set $U \subset \mathbb{R}^d$ and a diffeomorphism $\Phi:U \to O$,
    \begin{align*}
        \lambda(f, C) = \lambda(f \circ \Phi, \Phi^{-1}(C)),
    \end{align*}
    holds.
\end{proposition}

\begin{proof}
    Define the $\zeta$-function associated with $f$ and $C$ by
    \begin{align*}
        \zeta_1(s) &= \int_C f(w)^s dw, \\
        \zeta_2(s) &= \int_{\Phi^{-1} (C)} f(\Phi(u))^s du.
    \end{align*}
    It suffices to show that $\zeta_1$ and $\zeta_2$ have the same convergence/divergence behavior on the real axis.
    In what follows, let $s \in \mathbb{R}$.
    
    By a change of variables in the integral,
    \[
        \zeta_1(s) = \int_{\Phi^{-1} (C)} f(\Phi(u))^s |J_{\Phi}(u)| du,
    \]
    holds. 
    Since $\Phi$ is a diffeomorphism and $C$ is compact, $\Phi^{-1}(C)$ is also compact.
    Hence there exist constants $c, c' > 0$ such that $c \leq |J_{\Phi}(u)| \leq c'$ holds.
    Therefore,
    \[
        c \zeta_2(s) \leq \zeta_1(s) \leq c' \zeta_2(s)
    \]
    holds.
    Consequently, $\zeta_1$ and $\zeta_2$ have the same convergence/divergence behavior on the real axis, and thus we must have $\lambda(f, C) = \lambda( f \circ \Phi, \Phi^{-1}(C))$.
\end{proof}

Next, we show that the RLCT is determined only by a neighborhood of the zero set of the function.

\begin{proposition}[Removal of nonzero points]\label{prop:nonzero_ignore}
    Let $f$ be a nonnegative real analytic function $f: O \to \mathbb{R}$ defined on an open set $O \subset \mathbb{R}^d$ whose zero set $Z = \{x \in O : f(x)=0\}$ is compact.
	For any compact set $C \subset O$ containing $Z$, 
    we have 
    \[
    \lambda(f, C) = \lambda(f, T_{\delta}(Z)),
    \]
    for all sufficiently small $\delta > 0$ such that $T_{\delta}(Z) \subset C$, where $T_{\delta}(Z) = \{x \in O : d(x, Z) \le \delta\}$.
\end{proposition}

\begin{proof}
We have
\[
\zeta(s)
=
\int_{C} f(u)^s du
=
\int_{T_{\delta}(Z)} f^{s} du 
+
\int_{C \setminus T_{\delta}(Z)} f^{s} du.
\] 
Since the second integral has no poles, the poles of the zeta function are determined only by the first integral. Therefore, we obtain
\[
\lambda(f,C) = \lambda(f,T_{\delta}(Z)).
\]
\end{proof}

Finally, we compute the RLCT of a function $f$ whose Hessian is non-degenerate. 

\begin{proposition}[RLCT of a sum of squares]\label{lem:rlct_of_squre}
Let $f : \mathbb{R}^d \to \mathbb{R}$ be given by
\[
f(u)=\|u\|^2=u_1^2+\cdots+u_d^2.
\]
and $C$ contains the origin in its interior.
Then we have
\[
 \lambda(f, C) = \frac{d}{2}.
\]
\end{proposition}

\begin{proof}
Since $f$ vanishes only at the origin, for any \(\delta>0\) such that
\(B_{\delta}(0)\subset C\), we have $\lambda(f,C)=\lambda(f,B_{\delta}(0))$
 (Proposition~\ref{prop:nonzero_ignore}).
Consider the local zeta function near the origin:
\[
\zeta(s)
=
\int_{B_{\delta}(0)} f(u)^s du
=
\int_{B_{\delta}(0)} |u|^{2s} du.
\] 
Using polar coordinates $u = r\omega$ with $r \geq 0$ and $\omega \in S^{d-1}$, the $(d-1)$-dimensional unit sphere, we have $du = r^{d-1}\,dr\,d\omega$.
Therefore,
\[
\zeta(s)
=
\operatorname{vol}(S^{d-1})
\int_0^\delta r^{2s+d-1} dr = \operatorname{vol}(S^{d-1}) \frac{\delta^{2s + d}}{2s + d},
\]
where $\operatorname{vol}(S^{d-1})$ denotes the usual volume of $S^{d-1}$.
Since the rightmost pole on the right-hand side is $-\frac{d}{2}$, we have
\[
\lambda(f, C) = \frac{d}{2}.
\]
\end{proof}

\section{Computation of the Local RLCT} \label{appendix:local_rlct}

We explicitly compute the local RLCT claimed in Subsection~\ref{subsec:rrr} and \ref{subsec:lnn}.
The computational method differs depending on the point under consideration.

\subsection{Local RLCTs of Reduced Rank Regression}\label{appendix:local_rlct_RRR}

The computation in \eqref{eq:rrr-rlct} is as follows.

\subsubsection{Derivation of \(\lambda(w_1)\).}

Recall that $w_1 = (U_1, V_1)$ is given by 
\begin{equation*}\label{eq:def_w_rrr}
\begin{aligned}
    U_1 &=
        \begin{pmatrix}
            I_h \\
            0
        \end{pmatrix} \in \mathbb{R}^{m_2 \times h},
    \qquad
    V_1 =
        \begin{pmatrix}
            I_r & 0 \\
            0 & 0
        \end{pmatrix} \in \mathbb{R}^{h \times m_1},
\end{aligned}
\end{equation*}
Let $ U \in \mathbb R^{m_2\times h}$ and $V \in \mathbb R^{h\times m_1}$ be matrices in neighborhoods of $ U_1$ and $ V_1$, respectively.
Since $\mathbb{E}[XX^\top]$ is positive definite,
there exist constants $c,c'>0$ such that
\[
c\|UV-T^\star\|_F^2
\le R(w)
\le c'\|UV-T^\star\|_F^2.
\]
Hence, the local RLCT of $R$ at $w_1$ is equal to that of $\|UV-T^\star\|_F^2$ by Proposition~\ref{prop:muleq}.

Furthermore, we simplify the expression by applying a diffeomorphism.
Write
\[
U=\begin{pmatrix} \mathcal{U}_1 \\ \mathcal{U}_2 \end{pmatrix},
\qquad
T^\star=\begin{pmatrix} T_0\\0\end{pmatrix},
\]
where $\mathcal{U}_1 \in \mathbb R^{h\times h}$ and
\[
T_0=
\begin{pmatrix}
I_r&0\\
0&0
\end{pmatrix}
\in\mathbb R^{h\times m_1}.
\]
Since $\mathcal{U}_1=I_h$ at $w_1$, the matrix $\mathcal{U}_1$ is invertible in a sufficiently small neighborhood of $w_1$.
We can therefore write
\[
\begin{aligned}
UV-T^\star
&=
\begin{pmatrix}
\mathcal{U}_1V-T_0\\
\mathcal{U}_2V
\end{pmatrix}
=
\begin{pmatrix}
\mathcal{U}_1V-T_0\\
\mathcal{U}_2\mathcal{U}_1^{-1}
(\mathcal{U}_1V-T_0)
+\mathcal{U}_2\mathcal{U}_1^{-1}T_0
\end{pmatrix}.
\end{aligned}
\]
Then we introduce the diffeomorphic coordinate transformation $\Phi:(U,V)\mapsto(\mathcal{U}_1,G,H)$, where
\[
G=\mathcal{U}_1V-T_0,
\qquad
H=\mathcal{U}_2\mathcal{U}_1^{-1}.
\]
Under this transformation,
\[
UV-T^\star
=
\begin{pmatrix}
G\\
HG+HT_0
\end{pmatrix}.
\]
Letting 
$R = \begin{pmatrix}
I_h&0\\
-H&I_{m_2-h}
\end{pmatrix},$ 
we have
$
UV-T^\star
=
R^{-1} 
\begin{pmatrix}
G\\
HT_0
\end{pmatrix}.
$
Since $\Phi$ is a diffeomorphism and left multiplication by $R^{-1}$ preserves the Frobenius norm up to constant factors, Propositions~\ref{lem:diffeomor_invarient} and~\ref{prop:muleq} imply that $\|UV-T^\star\|_F^2$ and $\|G\|_F^2+\|HT_0\|_F^2$ have the same local RLCT.
By Proposition~\ref{lem:rlct_of_squre}, we obtain 
\[
\lambda(w_1)=\frac{h m_1 + r(m_2 - h)}{2}.
\]

\subsubsection{Derivation of $\lambda(w_2)$.}
Recall that $w_2=(U_2, V_2)$ is
\begin{equation*}\label{eq:def_w_rrr}
\begin{aligned}
    U_2 &=
        \begin{pmatrix}
            I_r & 0 \\
            0 & 0
        \end{pmatrix} \in \mathbb{R}^{m_2 \times h},
    \qquad
    V_2 =
        \begin{pmatrix}
            I_h & 0
        \end{pmatrix} \in \mathbb{R}^{h \times m_1}.
\end{aligned}
\end{equation*}

Writing
\[
V=\begin{pmatrix}\mathcal{V}_1&\mathcal{V}_2\end{pmatrix},
\qquad
T^\star=\begin{pmatrix}T_0&0\end{pmatrix},
\]
where $\mathcal{V}_1\in\mathbb R^{h\times h}$ and
\[
T_0=
\begin{pmatrix}
I_r&0\\
0&0
\end{pmatrix}
\in\mathbb R^{m_2\times h},
\]
the proof is essentially the same as that for $\lambda(w_1)$.
We obtain
\[
\lambda(w_2)=\frac{h m_2 + r(m_1 - h)}{2}.
\]

\subsubsection{Derivation of \(\lambda(w_3)\).}
Recall that $w_3 = (U_3, V_3)$ is given by 
\[
U_3 =
    \begin{pmatrix}
        I_r & 0 \\
        0 & 0
    \end{pmatrix} \in \mathbb{R}^{m_2 \times h},
\qquad
V_3 =
    \begin{pmatrix}
        I_r & 0
    \end{pmatrix} \in \mathbb{R}^{h \times m_1}.
\]
The main idea is to use Main Theorem Case~(2) of \cite{aoyagi2005stochastic}.
By Lemma 5 (a) of \cite{aoyagi2005stochastic},
the local RLCT of $R$ is equal to that of $\|S\|^2+\|\mathcal{UV}\|_F^2$,
where $S\in\mathbb R^{D_r}, D_r=r(m_1+m_2-r)$, $\mathcal{U} \in \mathbb{R}^{(h-r) \times (m_1-r)}$ and $\mathcal{V} \in \mathbb{R}^{(m_2-r) \times (h-r)}$.
Since $\|(t\mathcal{U}) (t\mathcal{V})\|_F^2 = t^4 \|\mathcal{UV}\|_F^2$ for $t \in \mathbb{R}$,
the RLCT is equal to the local RLCT. 
Indeed, letting $\varepsilon > 0$ and $p$ be the number of parameters of $(\mathcal{U},\mathcal{V})$, 
\[
\int_{\|(\mathcal{U},\mathcal{V})\|<\varepsilon} \|(t\mathcal{U}) (t\mathcal{V})\|_F^2s \,d\mathcal{U}\,d\mathcal{V}
=
\varepsilon^{p+4s}
\int_{\|(A,B)\|<1} \|\mathcal{UV}\|_F^2s\,dA\,dB,
\]
and $\varepsilon^{p+4s}$ does not change the poles.
Thus, we can use Main Theorem Case~(2) of \cite{aoyagi2005stochastic} for the local RLCT of $\|\mathcal{UV}\|_F^2$, and we have
\[
\lambda(w_3) =
\frac{1}{2}
\left[ r(m_1+m_2-r) + 
\min_{\substack{0 \le j \le h-r, j \in \mathbb{Z}}} \left\{j(m_2-r)+(m_1-r-j)(h-r-j)\right\} \right].
\]

\subsection{Local RLCTs of Linear Neural Networks}\label{appendix:local_rlct_LNN}

The computation in \eqref{eq:lnn-rlct} is as follows.
The computational method differs depending on the point under consideration.

\subsubsection{Derivation of \(\lambda(w_1)\).}

Recall that
$
w_1
=
 \left(
I_{h_1\times m_1},
I_{h_2\times h_1},
0_{m_2\times h_2}
\right).
$
Write
\[
W_{\mathrm{in}}
=
I_{h_1\times m_1}+E,
\qquad
W_{\mathrm{hid}}
=
I_{h_2\times h_1}+F,
\qquad
W_{\mathrm{out}}=G,
\]
in the neighborhood of $w_1$ where $E\in\mathbb{R}^{h_1\times m_1}, F\in\mathbb{R}^{h_2\times h_1}, G\in\mathbb{R}^{m_2\times h_2}$.
Then $w_1$ corresponds to $E=F=G=0$, and
$
W_{\mathrm{out}}W_{\mathrm{hid}}W_{\mathrm{in}}
=
G
\left(I_{h_2\times h_1}+F\right)
\left(I_{h_1\times m_1}+E\right).
$

Then the generalization error is
$
R(w)
=
\mathbb{E}
\left[
\left|
G
\left(I_{h_2\times h_1}+F\right)
\left(I_{h_1\times m_1}+E\right)X
\right|^2
\right].
$
Let $M = \left(I_{h_2\times h_1}+F\right) \left(I_{h_1\times m_1}+E\right)$, 
and note that $M \mathbb{E}[XX^{\top}] M^{\top}$ is positive definite in a sufficiently small neighborhood of $(E,F)=(0,0)$. 
Therefore there exist constants $c, c' > 0$ such that
\[
c \|G\|_F^2
\le
R(w)
\le
c'\|G\|_F^2.
\]
Thus, by Proposition~\ref{prop:muleq}, $R(w)$ has the same local RLCT at $w_1$ as $\|G\|_F^2$.

Since $G\in\mathbb{R}^{m_2\times h_2}$ consists of $m_2h_2$ parameters, by Proposition~\ref{lem:rlct_of_squre} we have
\[
\lambda(w_1)=\frac{m_2h_2}{2}.
\]

\subsubsection{Derivation of \(\lambda(w_2)\).}

Recall that
$
w_2
=
\left(
I_{h_1\times m_1},
0_{h_2\times h_1},
I_{m_2\times h_2}
\right).
$
Write
\[
W_{\mathrm{in}}
=
I_{h_1\times m_1}+E,
\qquad
W_{\mathrm{hid}}=F,
\qquad
W_{\mathrm{out}}
=
I_{m_2\times h_2}+G,
\]
in the neighborhood of $w_2$, where
$E \in \mathbb{R}^{h_1\times m_1}$,
$F \in \mathbb{R}^{h_2\times h_1}$, and
$G \in \mathbb{R}^{m_2\times h_2}$.
Then $w_2$ corresponds to $E=F=G=0$, 
and the generalization error is
$
R(w)
=
\mathbb{E}
\left[
\left|
\left(I_{m_2\times h_2}+G\right)
F
\left(I_{h_1\times m_1}+E\right)X
\right|^2
\right].
$

First, we absorb the dependence on $G$ into a local coordinate
transformation.
Since $m_2\le h_2$, write
\[
I_{m_2\times h_2}+G
=
\begin{pmatrix}
I_{m_2}+G_1 & G_2
\end{pmatrix},
\qquad
F=
\begin{pmatrix}
F_1\\
F_2
\end{pmatrix},
\]
where 
$G_1 \in \mathbb{R}^{m_2\times m_2}, G_2 \in \mathbb{R}^{m_2\times (h_2 - m_2)}, F_1\in\mathbb{R}^{m_2\times h_1}, F_2\in\mathbb{R}^{(h_2 - m_2) \times h_1}$.
Then $\left(I_{m_2\times h_2}+G\right)F =(I_{m_2}+G_1)F_1+G_2F_2$.
Define
$
\widetilde F
:=
(I_{m_2}+G_1)F_1+G_2F_2
\in\mathbb{R}^{m_2\times h_1}.
$
Since  $I_{m_2}+G_1$ is invertible in a sufficiently small
neighborhood of $G_1=0$, the transformation
$(F_1,F_2,G) \mapsto (\widetilde F,F_2,G)$ is invertible.
Therefore,
\[
R(w)
=
\mathbb{E}
\left[
\left|
\widetilde F
\left(I_{h_1\times m_1}+E\right)X
\right|^2
\right].
\]
Moreover, $I_{h_1\times m_1}+E$ has full row rank $h_1$ in a sufficiently small neighborhood of $E=0$.
Therefore, since $\left(I_{h_1\times m_1}+E\right) \mathbb{E}[XX^{\top}] \left(I_{h_1\times m_1}+E\right)^\top$ is positive definite, there exist constants $c,c' > 0$ such that
\[
c\|\widetilde F\|_F^2
\le
R(E,F,G)
\le
c'\|\widetilde F\|_F^2.
\]
Thus, by Proposition~\ref{prop:muleq}, $R(w)$ has the same local RLCT at $w_2$ as $\|\widetilde F\|_F^2$.
Since
\(\widetilde F\in\mathbb{R}^{m_2\times h_1}\) consists of
$m_2h_1$ parameters, by Proposition~\ref{lem:rlct_of_squre} we have
\[
\lambda(w_2)=\frac{m_2h_1}{2}.
\]

\subsubsection{Derivation of \(\lambda(w_3)\).}

Recall that
$
w_3
=
\left(
0_{h_1\times m_1},
I_{h_2\times h_1},
I_{m_2\times h_2}
\right).
$
Write
\[
W_{\mathrm{in}} = E,
\qquad
W_{\mathrm{hid}} = I_{h_2\times h_1}+F,
\qquad
W_{\mathrm{out}} = I_{m_2\times h_2}+G,
\]
in the neighborhood of $w_3$, where
$E \in \mathbb{R}^{h_1\times m_1}$,
$F \in \mathbb{R}^{h_2\times h_1}$ and
$G\in\mathbb{R}^{m_2\times h_2}$.
Then $w_3$ corresponds to $E=F=G=0$, 
and the generalization error is
$
R(w)
=
\mathbb{E}
\left[
\left|
\left(I_{m_2\times h_2}+G\right)
\left(I_{h_2\times h_1}+F\right)
EX
\right|^2
\right].
$

Let
$M(F,G) =
\left(I_{m_2\times h_2}+G\right)
\left(I_{h_2\times h_1}+F\right)
\in\mathbb{R}^{m_2\times h_1}$.
Noting that $M(0,0)=I_{m_2\times h_1}$ and $m_2\le h_1$, let
\[
M(F,G)
=
\begin{pmatrix}
M_1(F,G) & M_2(F,G)
\end{pmatrix},
\qquad
E=
\begin{pmatrix}
E_1\\
E_2
\end{pmatrix},
\]
where $M_1(F,G)\in\mathbb{R}^{m_2\times m_2}, M_2(F,G)\in\mathbb{R}^{m_2\times(h_1-m_2)}$
and $E_1\in\mathbb{R}^{m_2\times m_1}, E_2\in\mathbb{R}^{(h_1-m_2)\times m_1}$.
Since $M_1(0,0)=I_{m_2}$, $M_1(F,G)$ is invertible in a sufficiently small neighborhood of $(F,G)=(0,0)$.
Accordingly, define
$ 
\widetilde E
:=
M_1(F,G)E_1+M_2(F,G)E_2
=
M(F,G)E
\in\mathbb{R}^{m_2\times m_1}.
$
Then the transformation $(E_1,E_2,F,G) \mapsto (\widetilde E,E_2,F,G)$ is invertible. 
Therefore, there exist constants $c, c' > 0$ such that
\[
c\|\widetilde E\|_F^2
\le
R(w)
\le
c'\|\widetilde E\|_F^2.
\]
Thus, by Proposition~\ref{prop:muleq}, $R(w)$ has the same local RLCT at $w_3$ as $\|\widetilde E\|_F^2$.
Since
$\widetilde E\in\mathbb{R}^{m_2\times m_1}$ consists of
$m_2m_1$ parameters, by Proposition~\ref{lem:rlct_of_squre} we have
\[
\lambda(w_3)=\frac{m_2m_1}{2}.
\]

\section{High-Dimensional Statistical Techniques}\label{appendix:high-dimension}

By using techniques from high-dimensional statistics, we can derive high-probability upper bounds for stochastic processes.
In this chapter, we introduce some basic propositions.

\begin{proposition} \label{prop:syutyu_ineq}
	Let $A \subset W$ and $B = \sup_{w \in A} \|\ell(\cdot, w)\|_{\infty}$.
	For any $\eta > 0$,
	\begin{align*}
		\mathbb{P}\left( \sup_{w \in A} |R(w) - R_S(w)| 
		 \geq  2 \mathrm{Rad}(A) + \eta \right) 
			\leq \exp \left(- \frac{n\eta^2}{2B^2} \right), 
	\end{align*}
	holds, where 
    \[
        \mathrm{Rad}(A)
        :=
        \mathbb{E}_{Z}
        \mathbb{E}_{\sigma}
        \left[
        \sup_{w \in A}
        \left|
        \frac{1}{n}
        \sum_{i=1}^n
        \sigma_i \ell(Z_i,w)
        \right|
        \right],
    \]
    and each component of $\sigma = (\sigma_1,\dots,\sigma_n)$ is an independent Rademacher random variable, i.e.,
    \[
    \mathbb{P}(\sigma_i=1)=\mathbb{P}(\sigma_i=-1)=\frac{1}{2}.
    \]
\end{proposition}

\begin{proof}
    This follows from the standard Rademacher complexity bound; see, e.g.,
    \cite[Theorem 4.10]{wainwright2019high}.
\end{proof}

\begin{proposition}[Massart's Lemma]\label{prop:massart}
    Let $\mathcal{H}$ be a finite subset of $\mathbb{R}^n$, and $D = \max_{h \in \mathcal{H}} \|h\|_2$, then
    \[
        \mathbb{E}_{\sigma}
        \left[
        \max_{h \in \mathcal{H}}
        \frac{1}{n}
        \sum_{i=1}^n \sigma_i h_i
        \right] 
        \le \frac{D}{n} \sqrt{2 \log|\mathcal{H}|},
    \]
    where $\sigma = (\sigma_1, \ldots, \sigma_n)$ is a vector of independent Rademacher variables and $h_i$ is the $i$-th element of $h \in \mathcal{H}$.
\end{proposition}

\begin{proof}
If \(D=0\) or \(|\mathcal{H}|=1\), the claim is immediate. 
Hence assume \(D>0\) and \(|\mathcal{H}|\ge 2\).
For any \(\lambda>0\), we have
\begin{align*}
\mathbb{E}_{\sigma}
\left[
\max_{h \in \mathcal{H}}\frac{1}{n}\sum_{i=1}^n \sigma_i h_i
\right]
&=
\frac{1}{\lambda n}
\mathbb{E}_{\sigma}
\left[
\log \exp\left(
\lambda \max_{h \in \mathcal{H}}\langle \sigma,h\rangle
\right)
\right] \\
&=
\frac{1}{\lambda n}
\mathbb{E}_{\sigma}
\left[
\log \max_{h \in \mathcal{H}}
\exp\left(\lambda\langle \sigma,h\rangle\right)
\right] \\
&\le
\frac{1}{\lambda n}
\mathbb{E}_{\sigma}
\left[
\log \sum_{h \in \mathcal{H}}
\exp\left(\lambda\langle \sigma,h\rangle\right)
\right] \\
&\le
\frac{1}{\lambda n}
\log \sum_{v\in \mathcal{H}}
\mathbb{E}_{\sigma}
\left[
\exp\left(\lambda\langle \sigma,h\rangle\right)
\right],
\end{align*}
where the last step follows from Jensen's inequality.

For each fixed \(h \in \mathcal{H}\),
\begin{align*}
\mathbb{E}_{\sigma}
\left[
\exp\left(\lambda \langle \sigma,h\rangle\right)
\right]
&=
\prod_{i=1}^n
\mathbb{E}_{\sigma_i}
\left[
\exp\left(\lambda\sigma_i h_i\right)
\right] \\
&\le
\prod_{i=1}^n
\exp\left(\frac{\lambda^2 v_i^2}{2}\right) \\
&=
\exp\left(\frac{\lambda^2\|v\|^2}{2}\right) \\
&\le
\exp\left(\frac{\lambda^2 D^2}{2}\right),
\end{align*}
where first inequality follows from the fact that Rademacher variables are sub-Gaussian.
Choosing $\lambda = \sqrt{2\log |\mathcal{H}|}/D$ and performing a straightforward calculation, we obtain
\[
\mathbb{E}_{\sigma}
\left[
\sup_{h \in \mathcal{H}}\frac{1}{n}\sum_{i=1}^n \sigma_i h_i
\right]
\le
\frac{D}{n}\sqrt{2\log |\mathcal{H}|}.
\]
\end{proof}

\begin{proposition}[Truncated Dudley Integral]\label{prop:dudley}
Let $\mathcal F$ be a function set.
Assume that \(0\in \mathcal F\), and put $B:=\sup_{f \in \mathcal F}\|f\|_\infty$.
Then, for every \(0<\alpha<2B\),
\[
    \mathrm{Rad}(\mathcal F)
    :=
    \mathbb{E}_{Z}
    \mathbb{E}_{\sigma}
    \left[
        \sup_{f \in \mathcal F}
        \left|
        \frac{1}{n}
        \sum_{i=1}^n
        \sigma_i f(Z_i)
        \right|
    \right]
    \le
    4\alpha
    +
    \frac{12}{\sqrt n}
    \int_\alpha^{2B}
    \sqrt{
        \log N\bigl(\mathcal F,L^2(P_n),\tau\bigr)
    } d\tau.
\]
\end{proposition}

\begin{proof}
Since \(0 \in \mathcal F\) and \(\|f\|_\infty\le B\) for all
\(f\in\mathcal F\), 
we have $\|f\|_{L^2(P_n)} \le 2B\). 
Thus the singleton \(\{0\}\) is a \(2B\)-net of \(\mathcal F\).

Let \(\{\delta_j\}_{j\ge 0}\) be a decreasing dyadic sequence with $\delta_0=2B, \delta_j=2^{-j}\delta_0$.
Choose \(K\) such that $\alpha < \delta_{K+1} \le 2\alpha$.
For each \(j=0,\ldots,K\), let \(T_j\) be a \(\delta_j\)-net of
\(\mathcal F\) with
$|T_j|=N\bigl(\mathcal F,L^2(P_n),\delta_j\bigr)$.
We take \(T_0=\{0\}\). For each \(f\in\mathcal F\), choose
\(\pi_j f\in T_j\) such that $\| f - \pi_j f \|_{L^2(P_n)} \le \delta_j$.
Then
\[
    f
    =
    \pi_0 f
    +
    \sum_{j=1}^K(\pi_j f-\pi_{j-1}f)
    +
    (f-\pi_K f).
\]
Since \(\pi_0 f=0\), we obtain
\begin{align*}
    \sup_{f\in\mathcal F} \left| \frac{1}{n} \sum_{i=1}^n \sigma_i f(Z_i) \right|
    &\le
    \sum_{j=1}^K
    \sup_{f\in\mathcal F}
    \left|\frac{1}{n} \sum_{i=1}^n \sigma_i \left( \pi_j f (Z_i)-\pi_{j-1}f (Z_i) \right) \right|
    +
    \sup_{f\in\mathcal F}
    \left|\frac{1}{n} \sum_{i=1}^n \sigma_i \left(f(Z_i)-\pi_K f(Z_i)\right)\right|\\
    &\le  
    \sum_{j=1}^K
    \sup_{f\in\mathcal F}
    \left|\frac{1}{n} \sum_{i=1}^n \sigma_i (\pi_j f(Z_i)-\pi_{j-1}f(Z_i))\right|
    + \delta_K.
\end{align*}
where we used Schwarz inequality in the last step.
Taking $\mathbb{E}_{\sigma}$ on both sides, we have
\begin{align*}
    \mathbb{E}_{\sigma} \left[ \sup_{f\in\mathcal F} \left| \frac{1}{n} \sum_{i=1}^n \sigma_i f(Z_i) \right|\right]
    \le  
    \sum_{j=1}^K
    \mathbb{E}_{\sigma} \left[ 
    \sup_{f\in\mathcal F}
    \left|\frac{1}{n} \sum_{i=1}^n \sigma_i (\pi_j f(Z_i)-\pi_{j-1}f(Z_i))\right| \right]
    + \delta_K.
\end{align*}
By Proposition~\ref{prop:massart}, we obtain
\begin{align*}
    \mathbb{E}_{\sigma} \left[ 
    \sup_{f\in\mathcal F}
    \left|\frac{1}{n} \sum_{i=1}^n \sigma_i (\pi_j f(Z_i)-\pi_{j-1}f(Z_i))\right| \right]
    &\le 
    \frac{\sup_{f \in \mathcal F} \| \pi_j f-\pi_{j-1}f\|_2}{n} \sqrt{2 \log |T_j||T_{j-1}|}\\
    &\le 
    \frac{2 \sup_{f \in \mathcal F} \| \pi_j f-\pi_{j-1}f\|_2}{n} \sqrt{\log |T_j|}.
\end{align*}
Moreover we have,
\begin{align*}
    \sup_{f \in \mathcal{F}_{R}} \| \pi_{j} f - \pi_{j-1}f\|_2
    &\le \sup_{f \in \mathcal{F}_{R}} \| \pi_{j} f -f \|_2 + \sup_{f \in \mathcal{F}_{R}}\|f - \pi_{j-1}f\|_2, \\
    &\le \sqrt{n} \delta_j + \sqrt{n} \delta_{j-1}\\
    &\le \sqrt{n} \delta_j + 2 \sqrt{n} \delta_{j}\\
    &\le 3 \sqrt{n} \delta_j.
\end{align*}
Thus
\begin{align*}
    \mathbb{E}_{\sigma} \left[
    \sup_{f\in\mathcal F}
    \left|\frac{1}{n} \sum_{i=1}^n \sigma_i (\pi_j f(Z_i)-\pi_{j-1}f(Z_i))\right| \right]
    &\le 
    \frac{6 \delta_j}{\sqrt{n}} \sqrt{\log |T_j|}.
\end{align*}
Since $N\bigl(\mathcal F,L^2(P_n),\tau\bigr)$ is nonincreasing in $\tau$,
we have
\begin{align*}
\int_{\delta_{j+1}}^{\delta_j}  
    \sqrt{
        \log N\bigl(\mathcal F,L^2(P_n),\tau \bigr)  
    } d\tau
    &\ge (\delta_j - \delta_{j+1}) \sqrt{\log N\bigl(\mathcal F,L^2(P_n),\delta_j \bigr)}\\
    &\ge \frac{\delta_j}{2} \sqrt{\log N\bigl(\mathcal F,L^2(P_n),\delta_j \bigr)}.
\end{align*}
Therefore,
\begin{align*}
    \mathbb E_\sigma 
    \left[ \sup_{f\in\mathcal F} \left| \frac{1}{n} \sum_{i=1}^n \sigma_i f(Z_i) \right| \right]
    &\le
    \delta_K
    +
    \frac{6}{\sqrt n}
    \sum_{j=1}^{K}
    \delta_j
    \sqrt{
        \log N\bigl(\mathcal F,L^2(P_n),\delta_j\bigr)
    }  \\
    &\le
    4\alpha
    +
    \frac{12}{\sqrt n}
    \int_\alpha^{2B}
    \sqrt{
        \log N\bigl(\mathcal F,L^2(P_n),\tau\bigr)
    } d\tau,
\end{align*}
noting that $\alpha < \delta_{K+1} \le 2\alpha$.
Taking expectation over $Z=(Z_1, \ldots, Z_n)$ gives the desired bound.
\end{proof}

\end{appendices}

\bibliography{sn-bibliography}

\end{document}